\PassOptionsToPackage{table,dvipsnames}{xcolor}
\documentclass[11pt]{article}

\usepackage{tikz}
\usepackage{cuted}
\usepackage[raggedrightboxes]{ragged2e}

\makeatletter
\let\LQM@kernelbegin\begin
\let\LQM@kernelend\end
\makeatother

\usepackage{arabtex} 
\usepackage{utf8}

\makeatletter
\let\begin\LQM@kernelbegin
\let\end\LQM@kernelend
\makeatother

\transtrue

\usepackage{amsmath} 

\usepackage[preprint]{acl}
\usepackage{lmodern}
\usepackage{latexsym}

\usepackage{tcolorbox}

\usepackage{booktabs} 
\usepackage[T1]{fontenc}

\usepackage[utf8]{inputenc}
\usepackage{tcolorbox}

\usepackage{microtype}

\usepackage{inconsolata}
\usepackage{booktabs}
\usepackage{tabularx}
\usepackage{multirow}

\definecolor{errGeneral}{named}{Cyan}
\definecolor{errNE}{named}{Cyan}
\definecolor{errMWE}{named}{Cyan}
\definecolor{errOmission}{named}{Cyan}

\definecolor{errWrongVar}{named}{Blue}

\definecolor{headerGray}{gray}{0.9}

\definecolor{errSub}{named}{Gray}
\usepackage{booktabs}
\usepackage{multirow}
\usepackage{makecell}
\usepackage{geometry}
\definecolor{catCyan}{rgb}{0.88,1,1}
\definecolor{catYellow}{rgb}{1,1,0.88}

\definecolor{catOrange}{rgb}{1,0.9,0.8}   
\definecolor{catRed}{rgb}{1,0.85,0.85}    
\definecolor{catGreen}{rgb}{0.85,1,0.85}  
\definecolor{catMagenta}{rgb}{1,0.85,1}

\usepackage{longtable}
\usepackage{booktabs}
\usepackage{ragged2e}
\usepackage{array}
\usepackage{multirow}
\usepackage[table,dvipsnames]{xcolor}

\newcommand{\hlerror}[1]{\setlength{\fboxsep}{1pt}\colorbox{yellow!40}{#1}}

\definecolor{TblHeaderBG}{HTML}{0B2D39}
\definecolor{TblHeaderFG}{HTML}{FFFFFF}
\definecolor{DirBG}{HTML}{EAF3F5}
\definecolor{DirFrame}{HTML}{B7CAD1}

\definecolor{GeminiC}{HTML}{0F8B8D}
\definecolor{GemmaC}{HTML}{6D4C97}
\definecolor{FanarC}{HTML}{C77D00}
\definecolor{CommandC}{HTML}{1F4E79}

\newcommand{\mcell}[2]{\parbox[t]{#1}{\RaggedRight #2}}

\newcommand{\DirAbbrev}[1]{%
  \ifnum\pdfstrcmp{#1}{ENG}=0 EN\else
  \ifnum\pdfstrcmp{#1}{EN}=0 EN\else
  \ifnum\pdfstrcmp{#1}{MOR}=0 MO\else
  \ifnum\pdfstrcmp{#1}{UAE}=0 UA\else
  \ifnum\pdfstrcmp{#1}{EGY}=0 EG\else
  \ifnum\pdfstrcmp{#1}{PAL}=0 PA\else
  \ifnum\pdfstrcmp{#1}{YEM}=0 YE\else
  \ifnum\pdfstrcmp{#1}{MAU}=0 MA\else
  \ifnum\pdfstrcmp{#1}{JOR}=0 JO\else
  #1%
  \fi\fi\fi\fi\fi\fi\fi\fi\fi
}

\newcommand{\DirBoxWidth}{1.05cm}

\newcommand{\DirPill}[2]{%
  \begingroup
  \setlength{\fboxsep}{1.0pt}%
  \setlength{\fboxrule}{0.25pt}%
  \fcolorbox{DirFrame}{DirBG}{%
    \makebox[\DirBoxWidth][c]{\textsf{\scriptsize \DirAbbrev{#1}\,$\rightarrow$\,\DirAbbrev{#2}}}%
  }%
  \endgroup
}

\newcommand{\ModelShort}[1]{%
  \ifnum\pdfstrcmp{#1}{Gemini-F}=0 Gemini\else
  \ifnum\pdfstrcmp{#1}{gemini-2.5-flash}=0 Gemini\else
  \ifnum\pdfstrcmp{#1}{gemini-2.5-pro}=0 Gemini\else
  \ifnum\pdfstrcmp{#1}{gemma-3}=0 Gemma\else
  \ifnum\pdfstrcmp{#1}{gemma-3-27b}=0 Gemma\else
  \ifnum\pdfstrcmp{#1}{Fanar-1-9B}=0 Fanar\else
  \ifnum\pdfstrcmp{#1}{c4ai-command-r7b}=0 Command-R\else
  \ifnum\pdfstrcmp{#1}{c4ai-command-a}=0 Command-A\else
  \ifnum\pdfstrcmp{#1}{c4ai-command-a-03}=0 Command-A\else
  #1%
  \fi\fi\fi\fi\fi\fi\fi\fi\fi
}

\newcommand{\ModelBadge}[3]{%
  \begingroup
  \setlength{\fboxsep}{1.2pt}%
  \setlength{\fboxrule}{0.3pt}%
  \fcolorbox{#1!55}{#1!10}{%
    \parbox[t]{\dimexpr\linewidth-2\fboxsep-2\fboxrule\relax}{%
      \raggedright
      \textcolor{#1}{\scriptsize #2}\hspace{0.35em}%
      \textsf{\scriptsize \ModelShort{#3}}%
    }%
  }%
  \endgroup
}

\usepackage{tabularx}                 
\usepackage{booktabs}                 
\usepackage{xspace}                   

\usepackage{booktabs}
\usepackage{tabularx}
\usepackage{graphicx} 
\usepackage{xspace}

\newcolumntype{Y}{>{\centering\arraybackslash}m{1.5em}}
\newcolumntype{Z}{>{\raggedright\arraybackslash}X}

\definecolor{GBlue}{HTML}{4285F4}
\definecolor{GBlueBG}{HTML}{E8F0FE} 
\definecolor{GPurple}{HTML}{8E24AA}
\definecolor{GPurpleBG}{HTML}{F3E5F5} 
\definecolor{GGreen}{HTML}{0F9D58}
\definecolor{GGreenBG}{HTML}{E6F4EA} 
\definecolor{GRed}{HTML}{DB4437}
\definecolor{GRedBG}{HTML}{FCE8E6}   

\usepackage{tcolorbox}
\tcbuselibrary{skins,breakable}
\usepackage{microtype}
\usepackage{ragged2e}

\definecolor{NEaccent}{HTML}{6B6FD6} 
\definecolor{NEbg}{HTML}{F7F7FD}     
\definecolor{NEframe}{HTML}{D9DBF3}  
\definecolor{NEbad}{HTML}{B22222}    
\definecolor{NEgood}{HTML}{1B7F3B}   

\newcommand{\bad}[1]{\colorbox{NEbad!10}{\textcolor{NEbad}{#1}}}

\usepackage{tabularx}
\usepackage{tcolorbox}
\tcbuselibrary{skins}

\definecolor{NEaccent}{HTML}{6B6FD6}
\definecolor{NEframe}{HTML}{D9DBF3}
\definecolor{NEbg}{HTML}{F7F7FD}
\definecolor{NEbad}{HTML}{B22222}
\definecolor{NEgood}{HTML}{1B7F3B}

\renewcommand{\bad}[1]{\colorbox{NEbad!10}{\textcolor{NEbad}{#1}}}

\newcommand{\NEStrip}[6]{%
\begin{tcolorbox}[
  enhanced,
  colback=white, colframe=NEframe, boxrule=0.45pt,
  arc=2mm, left=1.6mm, right=1.6mm, top=1.2mm, bottom=1.2mm,
  borderline west={1.1pt}{0pt}{NEaccent},
  title={\sffamily\bfseries #1},
  coltitle=black,
  boxed title style={
    colback=NEbg, colframe=NEframe, boxrule=0.35pt, arc=2mm,
    left=4pt, right=4pt, top=2pt, bottom=2pt
  }
]
{\scriptsize\sffamily
\begin{tabularx}{\linewidth}{@{}X r@{}}
\textcolor{black!55}{\DialectTag{#2}{#3}} & \textcolor{black!55}{\ModelTag{#4}} \\
\end{tabularx}
}
\vspace{0.7mm}

{\scriptsize
\textbf{Source:} \parbox[t]{0.85\linewidth}{\raggedleft #5} \\ \vfill
\textbf{System:} \parbox[t]{0.85\linewidth}{\raggedleft #6}
}
\end{tcolorbox}
}

\usepackage{color,soul} 
\usepackage{booktabs}

\usepackage{multirow}
\usepackage[LAE,T1]{fontenc}
\usepackage[arabic,french,english]{babel}
\usepackage[raggedrightboxes]{ragged2e}
\usepackage{todonotes}
\usepackage{lscape}
\usepackage{natbib}
\usepackage{longtable}
\usepackage{enumitem}
\usepackage{makecell}
\usepackage{colortbl}
\usepackage{arydshln}
\usepackage{soul}

\usepackage{scrtime}
\usepackage[nomargin]{fnpct}
\usepackage[utf8]{inputenc}
\usepackage[ arabic , english]{babel}
\usepackage{colortbl}
\usepackage{etoolbox}
\usepackage{dirtytalk}
\usepackage{array}
\usepackage{pmboxdraw}
\usepackage{bbding}
\usepackage{changepage}
\usepackage{natbib}
\usepackage{times}
\usepackage{tabularx}
\usepackage{latexsym}
\usepackage[utf8]{inputenc}
\usepackage{tcolorbox}
\usepackage[utf8]{inputenc}
\usepackage{comment}
\usepackage{lscape}
\usepackage{booktabs}
\usepackage[locale=FR]{siunitx}
\usepackage{booktabs}
\usepackage{textcomp}
\usepackage{appendix}
\usepackage{longtable}
\usepackage{booktabs}
\usepackage{caption}
\usepackage{array}
\usepackage{geometry}
\usepackage{multirow}
\usepackage{tabularx}
\usepackage{makecell}
\usepackage{colortbl}
\setcode{utf8}
\usepackage[utf8]{inputenc}
\usepackage{tabularx}
\usepackage{booktabs}
\usepackage{pdflscape}

 \usepackage{float}

\usepackage{booktabs}
\usepackage{tabularx}
\usepackage{array}
\usepackage{amssymb} 
\usepackage{tikz}

\definecolor{GBlue}{HTML}{1967D2}   
\definecolor{GPurple}{HTML}{673AB7} 
\definecolor{GRed}{HTML}{D93025}    
\definecolor{GGreen}{HTML}{1E8E3E}  

\definecolor{geminiBlueMain}{HTML}{1967D2}
\definecolor{geminiBlueLight}{HTML}{E8F0FE}
\definecolor{geminiBlueBar}{HTML}{1A73E8}

\definecolor{geminiPurpleMain}{HTML}{673AB7}
\definecolor{geminiPurpleLight}{HTML}{F3E5F5}
\definecolor{geminiPurpleBar}{HTML}{8E24AA}

\definecolor{scenBG}{HTML}{F8F9FA} 

\renewcommand{\arraystretch}{1.1} 

\usepackage{booktabs}
\usepackage{siunitx}

\definecolor{aclteal}{HTML}{008080}
\definecolor{barcolor}{HTML}{5F9EA0} 
\definecolor{subbarcolor}{HTML}{C0C0C0} 

\definecolor{errNE}{named}{Apricot}
\definecolor{errMWE}{named}{SkyBlue}
\definecolor{errOmission}{named}{Plum}
\definecolor{errGeneral}{named}{LimeGreen}
\definecolor{errWrongVar}{named}{Salmon}
\definecolor{errSub}{named}{Gray}

\usepackage{booktabs}
\usepackage{colortbl}
\usepackage{siunitx}

\definecolor{BestRow}{HTML}{E9F7EF}

\usepackage{booktabs}

\usepackage{siunitx}
\definecolor{olivegreen}{RGB}{107,142,35}
\definecolor{lightolivegreen}{RGB}{157,192,105}
\definecolor{customgreen}{rgb}{0.13, 0.55, 0.13}

\usepackage{pifont}
\definecolor{blue}{RGB}{0,0,255}
\definecolor{lightblue}{RGB}{0,216,230}

\definecolor{purple}{RGB}{153, 50, 204} 

\definecolor{blue}{RGB}{0,0,255}
\definecolor{lightblue}{RGB}{0,216,230}
\usepackage{times}
\usepackage{latexsym}

\usepackage[T1]{fontenc}

\usepackage[utf8]{inputenc}

\usepackage{microtype}

\usepackage{inconsolata}

\usepackage{graphicx}

\title{%
  \raisebox{-0.8ex}{%
    \includegraphics[height=2.2\fontcharht\font`\B]{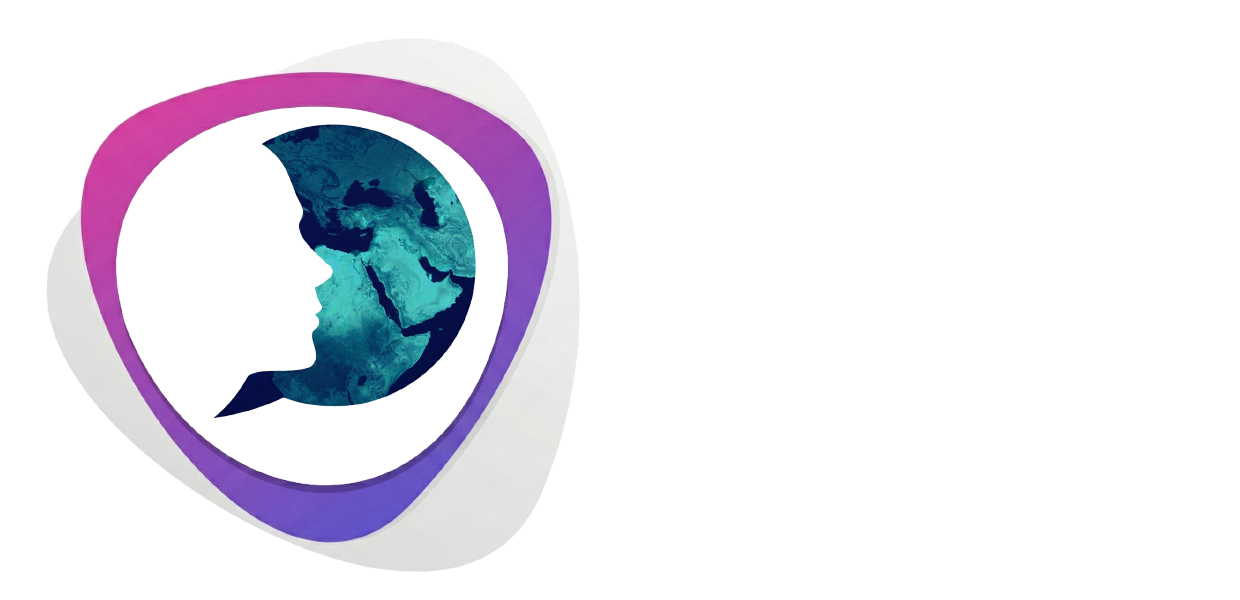}%
  }%
  \hspace{-1.2em}%
  LQM: Linguistically Motivated Multidimensional Quality Metrics for Machine Translation%
}

\author{
 \normalsize
\textbf{Samar {M. Magdy}}$^{\lambda}$ \quad
\textbf{Fakhraddin Alwajih}$^{\lambda}$\thanks{Equal contribution} \quad
\textbf{Abdellah {El Mekki}}$^{\lambda}$\footnotemark[1]\\
  \normalsize\textbf{Wesam {El Sayed}}$^{\xi}$ \quad
  \normalsize\textbf{Muhammad Abdul-Mageed}$^{\lambda,\gamma}$ \\
  \addlinespace
  $^{\lambda}$The University of British Columbia \quad $^{\xi}$Minia University \quad $^{\gamma}$Canada Research Chair in NLP and ML \\
  \texttt{\{samar.ahmad, muhammad.mageed\}@ubc.ca}
}

\usepackage{booktabs}
\usepackage{tabularx}
\usepackage{ragged2e}      

\definecolor{scen1blue}{HTML}{E3F2FD} 
\definecolor{scen1Head}{HTML}{1565C0} 

\definecolor{scen2cyan}{HTML}{E0F2F1} 
\definecolor{scen2Head}{HTML}{00695C} 

\definecolor{scen3green}{HTML}{E8F5E9} 
\definecolor{scen3Head}{HTML}{2E7D32} 

\definecolor{scen4red}{HTML}{FFEBEE}  
\definecolor{scen4Head}{HTML}{C62828} 

\newcolumntype{Y}{>{\centering\arraybackslash}X}
\setcode{utf8} 
\setarab 

\begin{document}
\maketitle

\begin{strip}
  \centering
  \includegraphics[width=\textwidth]{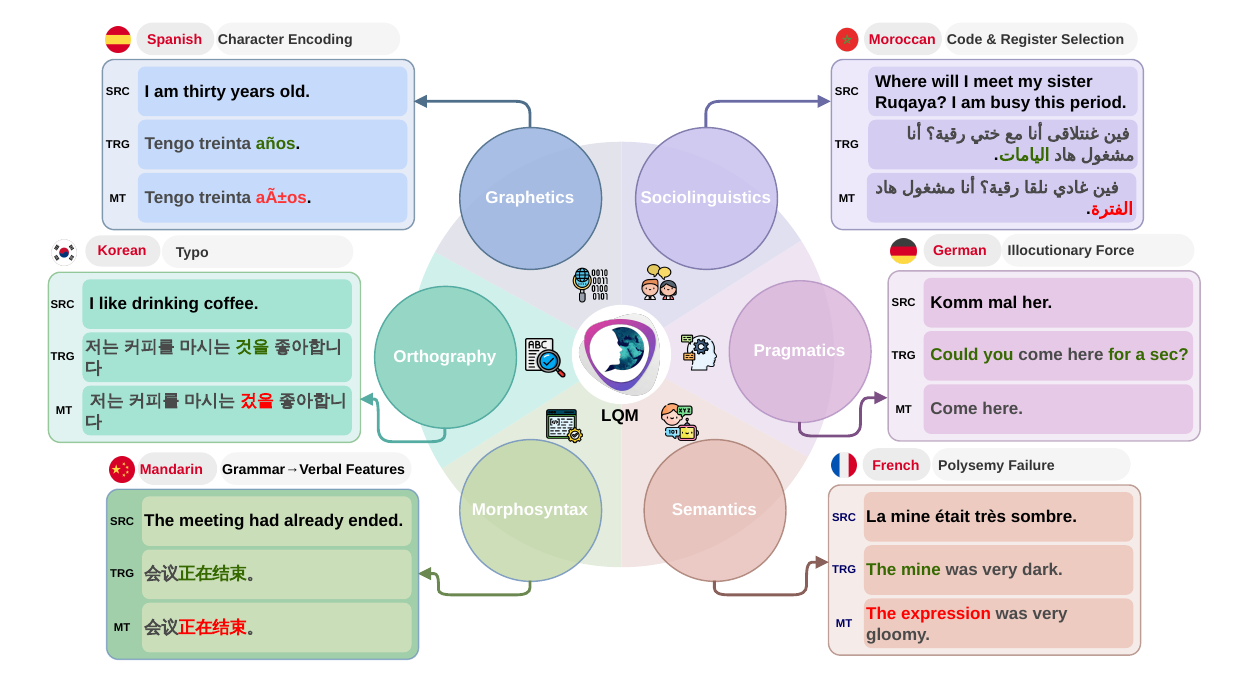}
  \captionof{figure}{Cross-lingual examples illustrating the proposed LQM framework's linguistic levels, demonstrating its language-agnostic design and broad applicability beyond Arabic.}
  \label{fig:lqm_main}
\end{strip}

\begin{abstract}\leavevmode
Existing MT evaluation frameworks, including automatic metrics and human evaluation schemes such as Multidimensional Quality Metrics (MQM), are largely language-agnostic. However, they often fail to capture dialect- and culture-specific errors in diglossic languages (e.g., Arabic), where translation failures stem from mismatches in language variety, content coverage, and pragmatic appropriateness rather than surface form alone.
We introduce \textbf{LQM}: Linguistically Motivated Multidimensional Quality Metrics for MT. LQM is a hierarchical error taxonomy for diagnosing MT errors through six linguistically grounded levels: \textit{sociolinguistics}, \textit{pragmatics}, \textit{semantics}, \textit{morphosyntax}, \textit{orthography}, and \textit{graphetics} (Figure~\ref{fig:lqm_main}).
We construct a bidirectional parallel corpus of $3{,}850$ sentences ($550$ per variety) spanning seven Arabic dialects (\textit{Egyptian}, \textit{Emirati}, \textit{Jordanian}, \textit{Mauritanian}, \textit{Moroccan}, \textit{Palestinian}, and \textit{Yemeni}), derived from conversational, culturally rich content. We evaluate six LLMs in a zero-shot setting and conduct expert span-level human annotation using LQM, producing 6{,}113 labeled error spans across 3{,}495 unique erroneous sentences, along with severity-weighted quality scores. We complement this analysis with an automatic metric (spBLEU). Though validated here on Arabic, LQM is a language-agnostic framework designed to be easily applied to or adapted for other languages. LQM annotated errors data, prompts, 
 and annotation guidelines are publicly available at \url{https://github.com/UBC-NLP/LQM_MT}.
\end{abstract}


\section{Introduction}
Evaluating MT remains challenging, particularly when systems must preserve sociolinguistic and pragmatic constraints in addition to semantic content. While automatic metrics such as BLEU~\cite{10.3115/1073083.1073135}, COMET~\cite{rei-etal-2020-comet}, and chrF++~\cite{popovic-2017-chrf} provide rapid, scalable feedback, they primarily capture surface overlap or embedding similarity and can diverge from human judgments in cases where translation quality depends on variety choice, register, or cultural appropriateness \cite{yao-etal-2024-benchmarking,chao2025natural}. This has motivated increased use of human-in-the-loop evaluation to characterize failure modes \cite{brewster2025evaluating}.

Among fine-grained human evaluation frameworks, MQM~\cite{Lommel2014MultidimensionalQM} has become a widely adopted standard, offering a flexible error taxonomy for identifying translation issues. However, in complex diglossic languages such as Arabic~\cite{ferguson1959diglossia, bassiouney2020arabic}, we find that MQM-style categorizations can under-specify dialect- and culture-conditioned errors. In particular, when errors reflect mismatches in dialect, register, or pragmatic intent, surface-oriented error labels may not provide enough structure to consistently localize the linguistic level at which a model fails.

In this work, we analyze the limitations that arise when applying MQM to bidirectional MT involving Arabic dialects, and  highlight three recurring challenges. \textit{(i)} MQM provides an operational inventory of error types, but it does not explicitly index errors to linguistic levels, which can make it difficult to separate similar surface manifestations with different underlying explanations. \textit{(ii)} The taxonomy offers limited built-in support for the systematic treatment of variation, including dialect choice, register, and culturally conditioned pragmatic constraints, which are central in dialectal Arabic. \textit{(iii)} As a result, MQM annotations can be harder to translate into targeted corrective signals for improving dialectal Arabic MT, since distinct phenomena (e.g., pragmatic infelicity vs.\ semantic reference errors) may be grouped under broad categories.

To address these gaps, we introduce Linguistically Motivated Multidimensional Quality Metrics, dubbed \textbf{LQM}, a linguistically grounded taxonomy designed to diagnose MT errors in a way that is aligned with linguistic theory and practical annotation. We develop LQM through a two-pronged process aimed at balancing theoretical coverage with empirical adequacy:
\textit{Top-down:} We applied MQM to a pilot subset of our MT data and analyzed where existing categories and guidelines were insufficient to consistently capture dialectal and sociopragmatic phenomena.
\textit{Bottom-up:} We then performed iterative, data-driven refinement over observed errors (span-level), consolidating recurring patterns into categories organized by linguistic level.
We synthesize these perspectives into a hierarchical framework spanning Sociolinguistics, Pragmatics, Semantics, Morphosyntax, Orthography, and Graphetics.
Our contributions are:
\textit{(i)} \textit{LQM framework:} We propose LQM, a linguistically grounded taxonomy that explicitly separates sociolinguistic and pragmatic phenomena from semantic and form-level errors, enabling more targeted diagnosis in diglossic and dialectal settings. The hierarchy supports two complementary annotation settings: a lightweight version for assigning broad error categories and a diagnostic version for labeling specific error types.
\textit{(ii)} \textit{Parallel dataset:} We introduce a conversational, culturally rich parallel corpus covering seven Arabic varieties to stress-test models under dialect-sensitive conditions.
\textit{(iii)} \textit{Multi-model evaluation:} We evaluate six LLMs with expert span-level human annotations (including severity-weighted scores) and complement this analysis with standard automatic metrics.

\section{Related works}
MQM has become a widely used framework for diagnostic MT evaluation and error taxonomy analysis \cite{Lommel2014MultidimensionalQM, Freitag2021ExpertsEA}. Recent work has pushed MQM-style evaluation toward finer granularity by localizing errors at the span level, including metric-based approaches such as xCOMET \cite{Guerreiro2023xcometTM} and prompted LLM evaluators such as GEMBA-MQM \cite{Kocmi2023GEMBAMQMDT, Fernandes2023TheDI}. Related directions include agentic or multi-step evaluators and refinement pipelines \cite{He2024ImprovingMT, wang2025drt}, as well as efforts to improve reliability by filtering or validating annotated spans via post-editing signals \cite{lu2025mqm, Kocmi2024ErrorSA, Kreutzer2020CorrectMI}.

In parallel, research in Automatic Post-Editing (APE) has evolved from classical formulations \cite{Simard2007StatisticalPP} to LLM-assisted pipelines \cite{Bhattacharyya2023FindingsOT, Raunak2023LeveragingGF}. Recent resources and protocols increasingly emphasize human-centered corrections and explainability, providing structured annotations or rationales that can support targeted diagnosis \cite{wasti2025translationcorrect, jung2024explainable, alves2024xtower}. Large Reasoning Models (LRMs) further enable multi-step explanations for ambiguity resolution \cite{Liu2025NewTF}, motivating work that constrains or structures LLM-based MQM annotation, for example, via compressed taxonomies (ThinMQM) or tagged span annotation \cite{Zhan2025AreLR, Yeom2025TaggedSA, He2025R1T1FI, wang2025deep, Feng2025MTR1ZeroAL}.

Despite these advances, applying general-purpose MQM-style frameworks to languages with diverse varieties such as Arabic remains challenging due to diglossia and dialectal variation \cite{ferguson1959diglossia, bassiouney2020arabic}, as well as non-standardized orthographies across dialects \cite{habash2012conventional}. Dialectal MT research has leveraged resources such as 
Alexandria \cite{mekki2026alexandria} and MADAR \cite{bouamor2018madar}, has increasingly studied LLM-based systems in specific dialectal settings \cite{yakhni2025lebanese, Fernandes2023TheDI}. Yet, morphological and syntactic variations continue to complicate error identification and interpretation \cite{zbib2012machine, sajjad-etal-2020-arabench}. Existing benchmarks such as Tarjamat \cite{kadaoui-etal-2023-tarjamat} and NADI 2024 \cite{abdul-mageed-etal-2024-nadi} underscore persistent gaps for low-resource dialects, but they do not provide a linguistically leveled, dialect-sensitive diagnostic taxonomy for isolating how and where translations lose dialectal identity or pragmatic appropriateness. 

These limitations motivate LQM, which organizes MQM-style error annotation by linguistic level (from sociolinguistics and pragmatics to form-level phenomena), enabling a more consistent diagnosis in dialectal Arabic settings.

\begin{figure*}[h]
\centering
\includegraphics[width=1.0\textwidth]{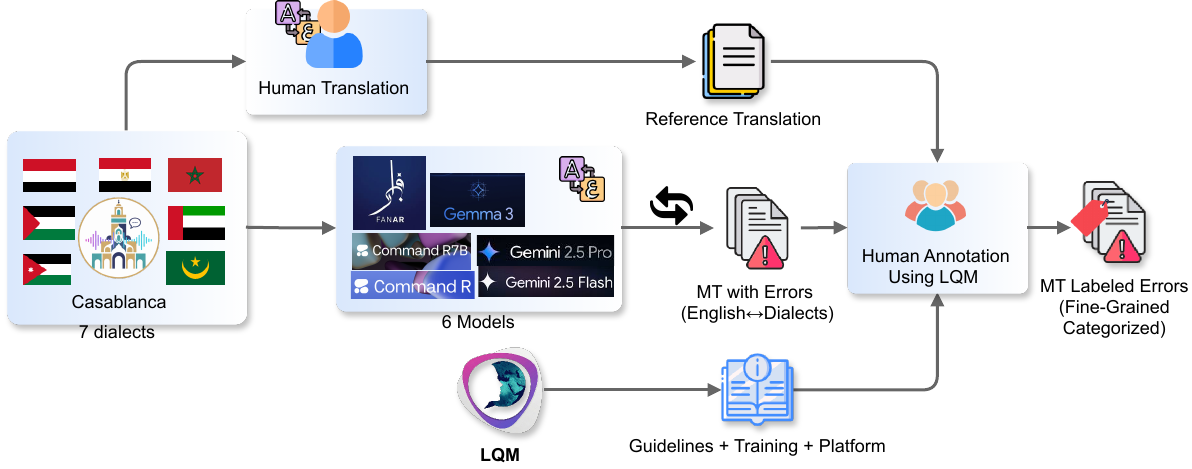}
\caption{Data and annotation workflow. Casablanca dataset of seven Arabic dialects is translated by humans and evaluated by six LLM models; creating LQM guidelines and training support human annotation to produce fine-grained MT error labels.}
\label{fig:lqm_workflow}
\end{figure*}

\section{Pitfalls of MQM}\label{sec:mqm_pitfalls}

MQM~\cite{Lommel2014MultidimensionalQM} is widely used for human diagnostic evaluation and is intentionally designed to be broadly applicable across languages. In our setting, bidirectional MT involving multiple Arabic dialects, we find that MQM's generic categories can under-specify error patterns that are primarily driven by variety choice and sociopragmatic constraints rather than surface form alone~\cite{10638285}.

\paragraph{Setup.} We applied MQM to translations from six Arabic-aware LLMs on a bidirectional translation dataset covering English and seven Arabic dialects. Six annotators—two senior linguists and four trained annotators—annotated a stratified sample of 50 sentences per model and translation direction. Each dialect was evaluated by annotators who are native speakers of that dialect. Full experimental details are provided in Section~\ref{sec:experiments}. 

\paragraph{Findings.} Table~\ref{tab:mqm_counts_updated} reports aggregated MQM span annotations.
A prominent pattern is the dominance of the \textit{Mistranslation} label ($60.09$\% of annotated errors). Based on qualitative inspection and annotator feedback, \textit{Mistranslation} frequently functions as a catch-all label for heterogeneous phenomena that do not fit cleanly under other MQM tags, reducing diagnostic specificity regarding where the model fails.
\begin{table}[t]
\centering
\footnotesize
\renewcommand{\arraystretch}{1.15}
\resizebox{\columnwidth}{!}{
\begin{tabular}{llrr}
\specialrule{1.5pt}{1pt}{1pt}
\rowcolor{gray!15} 
\textbf{MQM Category} & \textbf{MQM Subcategory} & \textbf{Count} & \textbf{Rate (\%)} \\
\midrule
\cellcolor{red!15}\textbf{Accuracy} & \cellcolor{red!15}\textbf{Mistranslation} & \cellcolor{red!15}\textbf{3{,}673} & \cellcolor{red!15}\textbf{60.09} \\
\rowcolor{white} Accuracy & Addition & 453 & 7.41 \\
\rowcolor{gray!6} Accuracy & Missing & 357 & 5.84 \\
\rowcolor{white} Fluency & Grammar & 325 & 5.32 \\
\rowcolor{gray!6} Style & Unidiomatic style & 322 & 5.27 \\
\rowcolor{white} Style & Awkward style & 292 & 4.78 \\
\rowcolor{gray!6} Terminology & Wrong term & 192 & 3.14 \\
\rowcolor{white} Accuracy & Undertranslation & 130 & 2.13 \\
\rowcolor{gray!6} Fluency & Spelling & 125 & 2.04 \\
\rowcolor{white} Accuracy & Overtranslation & 66 & 1.08 \\
\rowcolor{gray!6} Accuracy & Untranslated & 57 & 0.93 \\
\rowcolor{white} Terminology & Inconsistent with term resource & 47 & 0.77 \\
\rowcolor{gray!6} Style & Language register & 31 & 0.51 \\
\rowcolor{white} Style & Inconsistent style & 15 & 0.25 \\
\rowcolor{gray!6} Locale conv. & Currency format & 9 & 0.15 \\
\rowcolor{white} Fluency/Ling. conv. & Punctuation & 8 & 0.13 \\
\rowcolor{gray!6} Fluency & Inconsistency/unintelligible & 5 & 0.08 \\
\rowcolor{white} Fluency & Character encoding & 4 & 0.07 \\
\rowcolor{gray!6} Locale convention & Number format & 1 & < 0.02 \\
\rowcolor{white} Terminology & Inconsistent use of terminology & 1 & < 0.02 \\
\midrule
\rowcolor{gray!15} \textbf{Total} & & \textbf{6{,}113} & \textbf{100.00} \\
\specialrule{1.5pt}{1pt}{1pt}
\end{tabular}
}
\caption{MQM error counts and rates aggregated over annotated spans. The \colorbox{red!15}{Mistranslation} row is highlighted to indicate its frequent use as a catch-all label in our setting.}
\label{tab:mqm_counts_updated}
\end{table}

Annotators reported that many such cases reflect three recurring sources of error:
\textit{(i)} \textit{Pragmatic mismatches:} illocutionary force, discourse markers, vocatives, and honorifics.
\textit{(ii)} \textit{Variety consistency:} defaulting to MSA or drifting into a different dialect.
\textit{(iii)} \textit{Idiomatic usage:} mistranslation of proverbs and other fixed expressions.


\section{LQM Framework}
\label{sec:ara-mqm}
Motivated by the MQM analysis above, we introduce LQM, a linguistically grounded taxonomy for span-level human MT evaluation. LQM is designed to improve diagnostic precision in diglossic and dialectal settings by distinguishing sociolinguistic and pragmatic failures from semantic and form-level errors. Specifically, LQM organizes errors into six linguistic levels: \textit{Sociolinguistics}, \textit{Pragmatics}, \textit{Semantics}, \textit{Morphosyntax}, \textit{Orthography}, and \textit{Graphetics}.  We summarize these levels below; full definitions and additional examples are provided in Appendix~\S\ref{apendx:LQM framework}.

 \textbf{(i) Sociolinguistics.}
We place sociolinguistics at the top of the hierarchy to reflect communicative competence~\cite{hymes1972communicative}: a translation may be grammatically well-formed yet inappropriate for the social context. This is especially consequential in Arabic, where diglossia makes variety choice functional rather than merely stylistic~\cite{ferguson1959diglossia}. LQM therefore introduces \textit{code \& register selection} errors with three subcategories:
\textit{(a) standardization interference (vertical mismatch)},
\textit{(b) wrong dialect (horizontal mismatch)}, and
\textit{(c) register mismatch (tone/formality)}. 

\textbf{(ii) Pragmatics.}
While sociolinguistics targets broader social norms, the pragmatics level captures failures of communicative intent and implied meaning~\cite{levinson1983pragmatics}, i.e., mismatches between sentence meaning and speaker meaning. We group these under \textit{use, context, and cultural appropriateness} and include targeted subcategories:
\textit{(a) speech acts/illocutionary force},
\textit{(b) code switching},
\textit{(c) MWEs/proverbs},
\textit{(d) discourse marker mismatch}, and
\textit{(e) vocatives/honorifics /titles}~\cite{farwell1999pragmatics}.

\textbf{(iii) Semantics.}
This level evaluates meaning transfer, and preservation of propositional content~\cite{cruse1986lexical}. We distinguish:
\textit{(a) lexical semantics} (word meaning and lexical relations),
\textit{(b) propositional semantics} (truth-conditional content; avoiding unintended additions/omissions)~\cite{soames1987direct}, and
\textit{(c) discourse semantics} (cohesion and reference across sentences)~\cite{halliday2014cohesion, kamp2013discourse}.
Example subcategories include \textit{named entity}, \textit{wrong term}, \textit{polysemy failure}, and \textit{cross-variety interference}; see Appendix\S\ref{apendx:LQM framework} for complete definitions.

\textbf{(iv) Morphosyntax.}
This level captures violations of target-language structural constraints at the morphology-syntax interface~\cite{radford2004english}. We separate:
\textit{(a) grammar} (e.g., agreement and inflectional features)~\cite{corbett2006introduction}, including \textit{verbal features} (tense, aspect, voice, mood, person)~\cite{palmer2001mood} and \textit{nominal features} (number, gender, case, definiteness, state); and
\textit{(b) constituent order}~\cite{greenberg1963some}, including locale-sensitive reordering such as \textit{address format} and \textit{date format}.

\textbf{(v) Orthography/Writing Conventions.}
This level evaluates written-form conventions and mechanical correctness~\cite{derwing1992orthographic}. Since dialectal Arabic lacks fully standardized orthography, we accept dialectal spellings recognized by native annotators as conventional in informal digital contexts (e.g., texting and social media).\footnote{Given limited standardized dialect orthographies, we rely on annotator judgments for acceptability within emerging, socially shared informal conventions.}
We define five error types:
\textit{(a) spelling} (including \textit{typos/slips}),
\textit{(b) inconsistent spelling},
\textit{(c) unconventional Spelling},
\textit{(d) surface mechanics} (number, currency, time, telephone formats), and
\textit{(e) punctuation}.

\textbf{(vi) Graphetics.}
The lowest level captures failures in the technical realization of the text code. We include \textit{character encoding} errors, where the output is garbled due to encoding/decoding issues.

A comprehensive breakdown of LQM, including both the lightweight and diagnostic versions and their subcategories, is provided in  Table~\ref{tab:lqm} (Appendix). We also report an external validation of LQM conducted by two linguists specializing in linguistics and translation studies in Appendix~\S\ref{appendix:lqm_external_review}.

\section{Experimental Setup}\label{sec:experiments}

We conduct a case study to evaluate LQM in a stress-test setting for dialectal Arabic MT with Arabic-aware LLMs. Our experimental design mirrors the MQM analysis in Section~\ref{sec:mqm_pitfalls}, but replaces MQM with the LQM annotation protocol described in Section~\ref{sec:ara-mqm}. Below, we describe the dataset, the evaluated models, and our human and automatic evaluation procedures. Figure~\ref{fig:lqm_workflow} summarizes the data construction and LQM-based annotation workflow used in our experiments.

\subsection{Dataset}
\label{sec:dataset}
As part of our contribution, we construct a new bidirectional parallel corpus of $3{,}850$ sentences based on the manually transcribed conversational speech data introduced by \citet{talafha-etal-2024-casablanca}. The corpus covers seven Arabic varieties: \textit{Egyptian}, \textit{Emirati (UAE)}, \textit{Jordanian}, \textit{Mauritanian}, \textit{Moroccan}, \textit{Palestinian}, and \textit{Yemeni}. The dialectal transcripts were then professionally translated into MSA with support from native speakers of each dialect and subsequently translated into English.

The dataset is balanced with $550$ sentences per variety. It is organized for bidirectional MT: Dialect$\to$English (DA$\to$EN) and English$\to$Dialect (EN$\to$DA). We report DA$\to$EN results for all seven dialects. While for EN$\to$DA, the Jordanian, and Yemeni dialects were excluded due to the lack of available native-speaker translators for target-side validation. 

\subsection{Evaluated LLMs}
We select six Arabic-aware LLMs spanning both closed and open-weight systems. For closed-source models, we evaluate \texttt{Gemini-2.5-Pro} and \texttt{Gemini-2.5-Flash}~\cite{team2023gemini}. For open-weight models, we evaluate \texttt{Fanar-9B}~\cite{team2025fanar}, \texttt{Gemma-27B}~\cite{team2024gemma}, \texttt{Command-A (111B)}~\cite{cohere2024commandrplus}, and \texttt{Command-R7B (7B)}.
All models are evaluated in a zero-shot setting using a fixed prompt (Appendix\S\ref{appdx_fig:prompt_generator_no_msa_target}). 

\subsection{Human Labeling via LQM}

\paragraph{Annotation guidelines.}
LQM relies on expert human annotation to yield both \textit{(i)} fine-grained error diagnostics and \textit{(ii)} severity-weighted quality scores. We developed annotation guidelines that define LQM categories and provide Arabic examples to support consistent span selection and labeling. Comprehensive annotation guidelines are available in the previously mentioned repository.
   
\paragraph{Annotation scope.}
Annotators applied LQM to a random sample of $50$ translations per direction, drawn from the full set of outputs generated by the six models across dialects and directions. Across this annotated sample, annotators identified $3{,}495$ unique translations containing at least one error and labeled \textbf{$6{,}113$} error spans.\footnote{A sentence may receive multiple error labels, and the full set of $3{,}850$ sentence pairs also includes outputs judged error-free.}
Each labeled error includes: \textit{(i)} span boundaries, \textit{(ii)} LQM category/subcategory/sub-subcategory, \textit{(iii)} severity (minor/major/critical), and \textit{(iv)} an optional free-text explanation. Appendix~\S\ref{apendx:Data annotation} describes the annotation process, annotator profiles, and quality assurance workflow. Figure~\ref{fig:Examples_of_levels_of_LQM} presents representative examples from our dataset across different dialects, highlighting the error spans for each level. 

\renewcommand{\bad}[1]{\tcbox[
  on line,
  colback=red!18, colframe=red!18,
  boxsep=0.6pt, left=1pt, right=1pt, top=0.5pt, bottom=0.5pt,
  arc=0.8mm
]{#1}}

\newcommand{\ARHL}[1]{\tcbox[
  on line,
  colback=red!18, colframe=red!18,
  boxsep=0.6pt, left=1pt, right=1pt, top=0.5pt, bottom=0.5pt,
  arc=0.8mm
]{\AR{#1}}}

\newcommand{\VectorIcon}[2]{%
  \tikz[baseline=-0.8ex]{%
    \node[circle, fill=#2, inner sep=0pt, minimum size=1.2em] {%
      \color{white}\sffamily\bfseries\tiny #1%
    };%
  }%
}

\newcommand{\SeverityTag}[1]{%
  \smash{\raisebox{0.15ex}{%
    \tcbox[
      on line,
      colback=red!12,
      colframe=red!35,
      boxrule=0.35pt,
      boxsep=0.35pt,
      left=1.2pt, right=1.2pt, top=0.4pt, bottom=0.4pt,
      arc=0.7mm
    ]{\textcolor{red!75!black}{\sffamily\bfseries\tiny #1}}%
  }}%
}

\newcommand{\FineTypeTag}[1]{%
  \smash{\raisebox{0.12ex}{%
    \tcbox[
      on line,
      colback=NEaccent!10,
      colframe=NEaccent!25,
      boxrule=0.3pt,
      boxsep=0.35pt,
      left=1.2pt, right=1.2pt, top=0.35pt, bottom=0.35pt,
      arc=0.7mm
    ]{\textcolor{NEaccent}{\sffamily\bfseries\tiny #1}}%
  }}%
}
\renewcommand{\NEStrip}[8]{%
\begin{tcolorbox}[
    enhanced,
    colback=white,
    colframe=NEframe,
    boxrule=0.5pt,
    leftrule=2.5pt,
    arc=1mm,
    left=3pt, right=3pt, top=3pt, bottom=3pt,
    boxsep=1pt,
    borderline west={2.5pt}{0pt}{NEaccent},
    equal height group=LingLevels
]
    \noindent
    {\sffamily\bfseries\footnotesize\textcolor{NEaccent}{#1}%
    \makebox[0pt][l]{\hspace{0.35em}\raisebox{-1.45ex}{\FineTypeTag{#6}}}}%
    \hfill%
    {\scriptsize\textcolor{black}{%
        \llap{\SeverityTag{#5}\hspace{0.55em}}%
        \textbf{#2} \textcolor{black!40}{$\rightarrow$} \textbf{#3} \;\quad%
        \raisebox{-0.15em}{#4}%
    }}

    \vspace{2pt}

    {\footnotesize
    \begin{tabularx}{\linewidth}{@{}l X@{}}
       \textcolor{black!60}{\tiny\bfseries SRC} & #7 \\[-1pt]
       \textcolor{NEaccent}{\scriptsize$\hookrightarrow$} & #8 \\
    \end{tabularx}
    }
\end{tcolorbox}
}

\begin{figure*}[t]
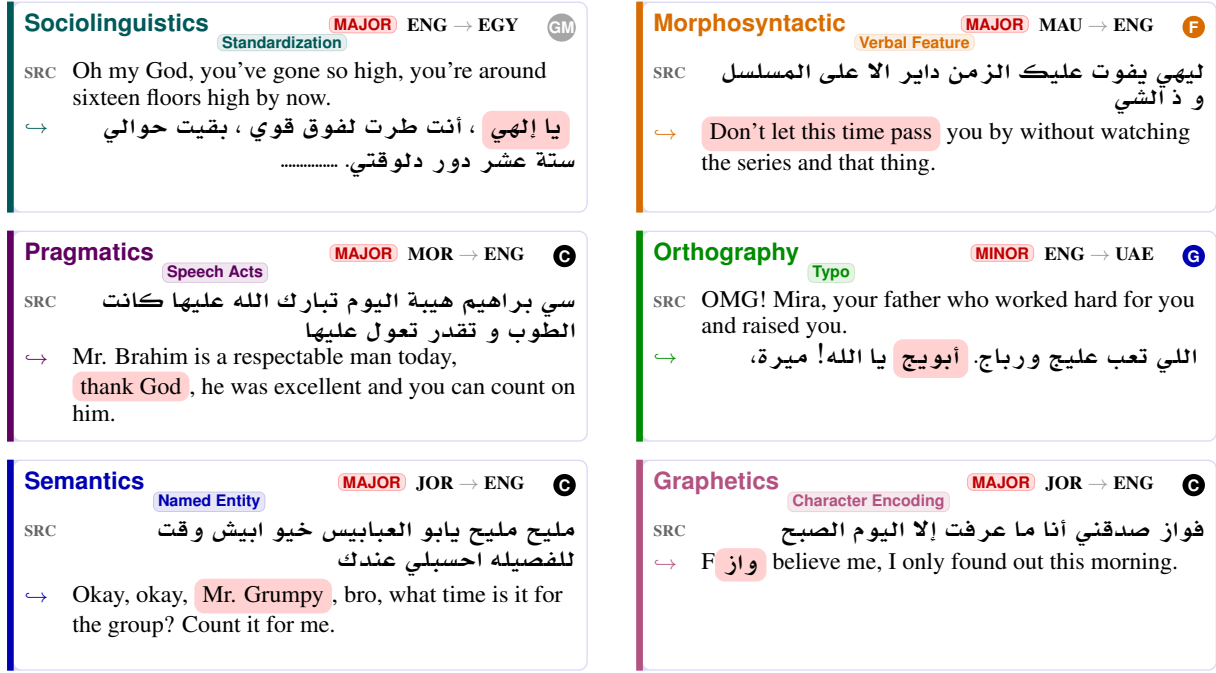

\centering
\setlength{\columnsep}{1.5em}

\begin{minipage}[t]{0.48\textwidth}

{\begingroup\colorlet{NEaccent}{teal!70!black}
\NEStrip{Sociolinguistics}
{ENG}{EGY}{\VectorIcon{GM}{gray!70}}{MAJOR}{Standardization}
{Oh my God, you’ve gone so high, you’re around sixteen floors high by now.}
{\raggedleft \AR{، أنت طرت لفوق قوي ، بقيت حوالي} \ARHL{يا إلهي} \AR{ ...............} \AR{ستة عشر دور دلوقتي.}}
\endgroup}

{\begingroup\colorlet{NEaccent}{violet!75!black}
\NEStrip{Pragmatics}
{MOR}{ENG}{\VectorIcon{C}{black}}{MAJOR}{Speech Acts}
{{\raggedleft \AR{سي براهيم هيبة اليوم تبارك الله عليها كانت الطوب و تقدر تعول عليها} \par}}
{Mr. Brahim is a respectable man today, \bad{thank God}, he was excellent and you can count on him.}
\endgroup}

{\begingroup\colorlet{NEaccent}{blue!70!black}
\NEStrip{Semantics}
{JOR}{ENG}{\VectorIcon{C}{black}}{MAJOR}{Named Entity}
{{\raggedleft \AR{مليح مليح يابو العبابيس خيو ابيش وقت للفصيله احسبلي عندك} \par}}
{Okay, okay, \bad{Mr. Grumpy}, bro, what time is it for the group? Count it for me.}
\endgroup}

\end{minipage}\hfill
\begin{minipage}[t]{0.48\textwidth}

{\begingroup\colorlet{NEaccent}{orange!85!black}
\NEStrip{Morphosyntactic}
{MAU}{ENG}{\VectorIcon{F}{orange!85!black}}{MAJOR}{Verbal Feature}
{{\raggedleft \AR{ليهي يفوت عليك الزمن داير الا على المسلسل و ذ الشي} \par}}
{\bad{Don't let this time pass} you by without watching the series and that thing.}
\endgroup}

{\begingroup\colorlet{NEaccent}{green!55!black}
\NEStrip{Orthography}
{ENG}{UAE}{\VectorIcon{G}{blue!70!black}}{MINOR}{Typo}
{OMG! Mira, your father who worked hard for you and raised you.}
{{\raggedleft \AR{يا الله! ميرة، } \ARHL{أبويج} \AR{ اللي تعب عليج ورباج.} \par}}
\endgroup}

{\begingroup\colorlet{NEaccent}{magenta!70!black}
\NEStrip{Graphetics}
{JOR}{ENG}{\VectorIcon{C}{black}}{MAJOR}{Character Encoding}
{{\raggedleft \AR{فواز صدقني أنا ما عرفت إلا اليوم الصبح} \par}}
{F\ARHL{واز} believe me, I only found out this morning.}
\endgroup}

\end{minipage}

\vspace{-2mm}
\caption{LQM-annotated examples across the six linguistic levels. Severity and fine-grained error types are indicated in each example. Error spans are highlighted in red. Icons indicate model families: \protect\VectorIcon{G}{blue!70!black} Gemini, \protect\VectorIcon{C}{black} Command, \protect\VectorIcon{F}{orange!85!black} Fanar, \protect\VectorIcon{GM}{gray!70} Gemma.}
\label{fig:Examples_of_levels_of_LQM}
\vspace{-3mm}
\end{figure*}


\subsection{Evaluation Metrics}
\subsubsection{LQM score derivation}
Our LQM score equation is inspired by the MQM original scoring equation ~\cite{Lommel2014MultidimensionalQM}, we follow the same severity weights $s_i$: \textit{minor}=1, \textit{major}=5, and \textit{critical}=25. Then, we fix a weight of $1$ across all the error types. For a translation with length $L$ (in words) and annotated errors $\{i\}$ with severity weights $s_i$, we compute:
\begin{equation*}
\text{LQM\_Score} = \max\left(0,\; 100\times\left(1 - \frac{\sum_i s_i}{L}  \right)\right)
\end{equation*}
This yields a normalized score in $[0,100]$, where higher values indicate better translation quality. LQM scores are based on human judgments (span selection and severity assignment) rather than automatic matching metrics.
\subsubsection{Automatic surface metrics}
We report automatic scores using \textit{spBLEU}~\cite{goyal-etal-2022-flores}, a sentence-level variant of BLEU~\cite{10.3115/1073083.1073135} to compute the correlation between LQM human annotated errors and automatic scores. We do not report model-based metrics such as COMET~\cite{rei-etal-2020-comet}, since their performance and calibration may be less reliable for dialectal Arabic varieties that are underrepresented in common training and evaluation resources.

\begin{table*}[!ht]
\centering
\footnotesize
\renewcommand{\arraystretch}{1.15}
\rowcolors{2}{gray!6}{white}
\resizebox{\textwidth}{!}{
\begin{tabular}{lllrr}
\specialrule{1.5pt}{1pt}{1pt}
\rowcolor{gray!15} 
\textbf{LQM Category} & \textbf{LQM Subcategory} & \textbf{LQM Subsubcategory} & \textbf{Count} & \textbf{Rate (\%)} \\
\midrule
sociolinguistics & code \& register selection & standardization interference (vertical mismatch) & 904 & 14.8 \\
semantics & lexical semantics & named entity & 572 & 9.4 \\
sociolinguistics & code \& register selection & wrong dialect (horizontal mismatch) & 539 & 8.8 \\
semantics & lexical semantics & coverage: unknown term/dialect & 365 & 6.0 \\
semantics & propositional semantics & omission & 357 & 5.8 \\
semantics & lexical semantics & unnatural/ unidiomatic style & 322 & 5.3 \\
morphosyntax & grammar & verbal features & 295 & 4.8 \\
semantics & lexical semantics & awkward style & 292 & 4.8 \\
pragmatics & use, context, cultural appropriateness & mwes, proverbs \& metaphors & 289 & 4.7 \\
semantics & discourse semantics & pronouns & 268 & 4.4 \\
semantics & propositional semantics & addition & 263 & 4.3 \\
semantics & lexical semantics & disambiguation: cross-variety interference & 246 & 4.0 \\
semantics & lexical semantics & wrong term & 192 & 3.1 \\
semantics & propositional semantics & hallucination & 190 & 3.1 \\
semantics & lexical semantics & undertranslation & 130 & 2.1 \\
orthography/ writing conventions & spelling & typo / slip & 125 & 2.0 \\
semantics & lexical semantics & disambiguation: polysemy failure & 120 & 2.0 \\
semantics & lexical semantics & transliteration & 115 & 1.9 \\
pragmatics & use, context, cultural appropriateness & speech acts mismatch & 104 & 1.7 \\
pragmatics & use, context, cultural appropriateness & forms of address (vocatives/honorifics/titles) & 97 & 1.6 \\
semantics & lexical semantics & overtranslated & 66 & 1.1 \\
semantics & lexical semantics & untranslated & 57 & 0.9 \\
semantics & discourse semantics & inconsistent with terminology resource & 47 & 0.8 \\
pragmatics & use, context, cultural appropriateness & discourse marker mismatch & 35 & 0.6 \\
sociolinguistics & code \& register selection & register mismatch & 31 & 0.5 \\
morphosyntax & grammar & nominal features & 30 & 0.5 \\
pragmatics & use, context, cultural appropriateness & code switching & 19 & 0.3 \\
semantics & discourse semantics & inconsistent style & 15 & 0.2 \\
orthography/ writing conventions & surface mechanics & currency format & 9 & 0.1 \\
orthography/ writing conventions & punctuation & --- & 8 & 0.1 \\
semantics & lexical semantics & unintelligible & 5 & <0.1 \\
graphetics & character encoding & --- & 4 & <0.1 \\
orthography/ writing conventions & surface mechanics & number format & 1 & <0.1 \\
semantics & discourse semantics & inconsistent use of terminology & 1 & <0.1 \\
\midrule
\rowcolor{gray!15} \textbf{Total} & & & \textbf{6{,}113} & \textbf{100.0} \\
\specialrule{1.5pt}{1pt}{1pt}
\end{tabular}
}
\caption{LQM error counts and rates utilizing the LQM Diagnostic layer for fine-grained analysis.}
\label{tab:lqm_counts_updated}
\end{table*}

\begin{table*}[t]
\centering
\newcommand{\win}[1]{\cellcolor{NEgood!15}\textbf{#1}}
\newcommand{\Dir}[2]{\textsc{#1}$\to$\textsc{#2}}
\footnotesize
\setlength{\tabcolsep}{0pt}
\renewcommand{\arraystretch}{1.1}
\begin{tabular*}{\textwidth}{@{\extracolsep{\fill}} l cccccc }
\toprule
\textbf{Direction} & 
\textbf{Fanar-9B} & 
\textbf{Command-A} & 
\textbf{Command-R7B} & 
\textbf{Gemma-27b} & 
\textbf{Gemini-2.5-flash} & 
\textbf{Gemini-2.5-pro} \\
\midrule
\rowcolor{NEbg} \multicolumn{7}{c}{\textsc{English $\to$ Dialect}} \\
\midrule[0.1pt]
\Dir{Eng}{Egy} & 49.90 & 71.14 & 35.89 & 54.44 & 65.23 & \win{72.31} \\
\Dir{Eng}{Mau} & \win{45.26} & 41.87 & 37.40 & 19.62 & 38.28 & 40.90 \\
\Dir{Eng}{Mor} & 6.46 & 38.60 & 17.00 & 19.29 & 51.53 & \win{68.60} \\
\Dir{Eng}{Pal} & 38.96 & 59.89 & 43.29 & 51.91 & 66.56 & \win{67.14} \\
\Dir{Eng}{Uae} & 78.32 & 72.89 & 21.94 & 63.39 & 82.10 & \win{83.07} \\
\addlinespace[4pt]
\rowcolor{NEbg} \multicolumn{7}{c}{\textsc{Dialect $\to$ English}} \\
\midrule[0.1pt]
\Dir{Egy}{Eng} & 53.57 & 60.88 & 45.56 & 68.69 & 74.45 & \win{75.89} \\
\Dir{Jor}{Eng} & 65.27 & \win{73.26} & 58.22 & 69.76 & 72.27 & 66.19 \\
\Dir{Mau}{Eng} & 40.76 & 56.38 & 43.90 & 61.45 & 59.26 & \win{63.88} \\
\Dir{Mor}{Eng} & 40.98 & 64.45 & 51.50 & 62.34 & \win{72.15} & 70.32 \\
\Dir{Pal}{Eng} & 64.78 & 73.18 & 62.62 & 72.89 & 73.42 & \win{79.47} \\
\Dir{Uae}{Eng} & 43.85 & 66.61 & 45.08 & 54.06 & 62.65 & \win{67.09} \\
\Dir{Yem}{Eng} & 61.28 & 62.90 & 58.07 & 67.00 & 70.76 & \win{73.41} \\
\bottomrule
\end{tabular*}
\caption{LQM scores (severity-weighted, 0--100) by model and direction. Higher is better.}
\label{tab:lqm_scores}
\end{table*}



\definecolor{HeaderTint}{HTML}{E9EEF5}   
\definecolor{SectionTint}{HTML}{F4F7FB}  
\definecolor{BestCell}{HTML}{DDF3E4}     

\begin{table*}[t]
\centering
\newcommand{\best}[1]{\cellcolor{BestCell}\textbf{#1}}
\newcommand{\Dir}[2]{\textsc{#1}$\,\rightarrow\,$\textsc{#2}}

\footnotesize
\setlength{\tabcolsep}{4pt}      
\renewcommand{\arraystretch}{1.18}

\begin{tabular*}{\textwidth}{@{\extracolsep{\fill}}lcccccc@{}}
\toprule

\rowcolor{HeaderTint}
& \multicolumn{4}{c}{\textbf{Open-Weight Models}} 
& \multicolumn{2}{c}{\textbf{Proprietary Models}} \\
\cmidrule(lr{6pt}){2-5} \cmidrule(l){6-7}

\textbf{Direction} & 
\textbf{Fanar-9B} & 
\textbf{Command-A} & 
\textbf{Command-R7B} & 
\textbf{Gemma-27B} & 
\textbf{Gemini-2.5-Flash} & 
\textbf{Gemini-2.5-Pro} \\
\midrule

\rowcolor{SectionTint}
\multicolumn{7}{c}{\textbf{English $\rightarrow$ Dialect}} \\
\specialrule{0.3pt}{0pt}{0pt}
\Dir{Eng}{Egy} & 13.97 & 23.38 & 13.15 & 19.84 & 24.11 & \best{26.09} \\
\Dir{Eng}{Mau} & 1.42 & 1.75 & 2.61 & 4.24 & 4.34 & \best{5.88} \\
\Dir{Eng}{Mor} & 2.64 & 10.18 & 6.97 & 9.28 & 15.39 & \best{18.30} \\
\Dir{Eng}{Pal} & 9.65 & 15.84 & 12.20 & 16.48 & 21.07 & \best{23.26} \\
\Dir{Eng}{Uae} & 5.67 & 13.21 & 5.89 & 11.17 & 17.78 & \best{19.77} \\

\addlinespace[0.45em]

\rowcolor{SectionTint}
\multicolumn{7}{c}{\textbf{Dialect $\rightarrow$ English}} \\
\specialrule{0.3pt}{0pt}{0pt}
\Dir{Egy}{Eng} & 28.89 & 31.62 & 26.70 & 27.54 & \best{32.18} & 31.47 \\
\Dir{Jor}{Eng} & 29.19 & 31.78 & 26.56 & 28.99 & \best{32.22} & 31.88 \\
\Dir{Mau}{Eng} & 10.13 & 12.82 & 8.96 & 11.19 & 16.03 & \best{16.59} \\
\Dir{Mor}{Eng} & 16.99 & 23.19 & 17.64 & 19.27 & \best{24.00} & 23.34 \\
\Dir{Pal}{Eng} & 25.67 & \best{31.17} & 23.47 & 25.87 & 30.56 & 27.12 \\
\Dir{Uae}{Eng} & 20.91 & 27.18 & 19.83 & 23.36 & \best{27.89} & 26.90 \\
\Dir{Yem}{Eng} & 22.54 & 23.98 & 20.17 & 22.09 & 25.97 & \best{26.26} \\

\bottomrule
\end{tabular*}

\caption{\textbf{spBLEU scores} across translation directions and model families. Higher is better. The best result in each row is highlighted in green and boldfaced.}
\label{tab:spbleu_scores}
\end{table*}
\begin{table*}[t]
\centering
\small
\setlength{\tabcolsep}{4pt}
\renewcommand{\arraystretch}{1.2}
\definecolor{badred}{HTML}{F8D7DA}
\definecolor{goodteal}{HTML}{D1E7DD}
\definecolor{rowgray}{gray}{0.96}
\definecolor{headergray}{gray}{0.90}
\begin{tabular}{l | cccccc | cccccc}
\toprule
\multirow{2}{*}{\textbf{Direction}} & 
\multicolumn{6}{c|}{\textbf{PART I: Model Error Contribution (\%)} \textit{(Who Failed?)}} & 
\multicolumn{6}{c}{\textbf{PART II: Error Type Distribution (\%)} \textit{(Why?)}} \\
\cmidrule(r){2-7} \cmidrule(l){8-13}
& \textbf{Cmd-A} & \textbf{Cmd-R} & \textbf{Fanar} & \textbf{Gemma} & \textbf{Flash} & \textbf{Pro} 
& \textbf{Soc} & \textbf{Prag} & \textbf{Sem} & \textbf{Morph} & \textbf{Orth} & \textbf{Gra} \\
\midrule
\rowcolor{headergray}
\multicolumn{13}{c}{\textbf{\textit{Dialect ~$\to$~ English}}} \\
\rowcolor{rowgray}
EGY~$\to$~ENG & 8.0 & \cellcolor{badred}\textbf{26.7} & 26.6 & 14.8 & 12.2 & 11.8 & 2.4 & 24.3 & \cellcolor{goodteal}\textbf{67.7} & 5.4 & 0.2 & -- \\
JOR~$\to$~ENG & 12.3 & \cellcolor{badred}\textbf{26.2} & 8.9 & 25.5 & 18.6 & 8.5 & 0.4 & 14.8 & \cellcolor{goodteal}\textbf{82.6} & 1.6 & 0.4 & 0.2 \\
\rowcolor{rowgray}
MAU~$\to$~ENG & 18.1 & 18.1 & \cellcolor{badred}\textbf{20.0} & 19.5 & 14.0 & 10.3 & 0.7 & 3.4 & \cellcolor{goodteal}\textbf{92.3} & 3.4 & 0.2 & -- \\
MOR~$\to$~ENG & 15.5 & 21.6 & \cellcolor{badred}\textbf{23.3} & 19.2 & 8.4 & 12.0 & -- & 10.4 & \cellcolor{goodteal}\textbf{85.1} & 3.9 & 0.6 & -- \\
\rowcolor{rowgray}
PAL~$\to$~ENG & 15.0 & \cellcolor{badred}\textbf{23.3} & 18.4 & 18.8 & 15.0 & 9.5 & -- & 10.9 & \cellcolor{goodteal}\textbf{87.0} & 1.4 & 0.6 & -- \\
UAE~$\to$~ENG & 13.4 & \cellcolor{badred}\textbf{23.8} & \cellcolor{badred}\textbf{23.8} & 14.9 & 13.0 & 11.2 & -- & 10.8 & \cellcolor{goodteal}\textbf{84.5} & 4.1 & 0.4 & 0.2 \\
\rowcolor{rowgray}
YEM~$\to$~ENG & 16.3 & \cellcolor{badred}\textbf{18.8} & 18.6 & 13.5 & 17.4 & 15.4 & 0.7 & 11.9 & \cellcolor{goodteal}\textbf{84.4} & 2.8 & 0.2 & -- \\
\midrule
\rowcolor{headergray}
\multicolumn{13}{c}{\textbf{\textit{English ~$\to$~ Dialect}}} \\
\rowcolor{rowgray}
ENG~$\to$~EGY & 19.7 & \cellcolor{badred}\textbf{30.8} & 16.8 & 14.4 & 11.3 & 7.0 & \cellcolor{badred}\textbf{40.1} & 7.2 & 39.9 & 8.4 & 4.3 & 0.2 \\
ENG~$\to$~MAU & 12.5 & 16.2 & 18.3 & \cellcolor{badred}\textbf{22.7} & 16.7 & 13.7 & \cellcolor{badred}\textbf{81.5} & 4.9 & 11.8 & 1.9 & -- & -- \\
\rowcolor{rowgray}
ENG~$\to$~MOR & 14.4 & 20.4 & \cellcolor{badred}\textbf{33.0} & 23.9 & 6.2 & 2.1 & \cellcolor{badred}\textbf{39.6} & 0.6 & 35.9 & 19.2 & 4.7 & -- \\
ENG~$\to$~PAL & 15.9 & 22.5 & \cellcolor{badred}\textbf{24.7} & 18.3 & 9.5 & 9.1 & \cellcolor{badred}\textbf{58.5} & 2.4 & 20.5 & 6.0 & 12.4 & 0.2 \\
\rowcolor{rowgray}
ENG~$\to$~UAE & 15.1 & \cellcolor{badred}\textbf{34.4} & 17.3 & 13.8 & 10.0 & 9.3 & \cellcolor{badred}\textbf{70.6} & 4.5 & 15.4 & 5.0 & 4.5 & -- \\
\bottomrule
\end{tabular}
\caption{\textbf{Diagnostic dashboard of dialectal failures.} Model error contribution (Part I) and error-type distribution (Part II) by translation direction. In \textit{DA$\rightarrow$EN} (top), \textbf{Semantic} errors dominate across dialects. In \textit{EN$\rightarrow$DA} (bottom), \textbf{Sociolinguistic} errors dominate overall, especially for \textbf{Mauritanian}, \textbf{Palestinian}, and \textbf{Emirati}, while \textbf{Egyptian} and \textbf{Moroccan} show a more mixed pattern, with \textbf{Semantic} errors remaining prominent and \textbf{Morphosyntax} also notable in Moroccan. \textit{(Soc=Sociolinguistics, Prag=Pragmatics, Sem=Semantics, Morph=Morphosyntax, Orth=Orthography, Gra=Graphetics).}}
\label{Error-distibutions-per-category+M}
\end{table*}
\section{\texorpdfstring{Results of Error Analysis}{Results of Error Analysis}}
\subsection{LQM Error Counts and Rates}
Table~\ref{tab:lqm_counts_updated} summarizes aggregated error counts and rates under LQM. Compared to MQM, which frequently assigned heterogeneous phenomena to the broad \textit{Mistranslation} label, LQM reallocates these cases into linguistically interpretable categories. In particular, many instances previously collapsed under the MQM \textit{Mistranslation} are separated into \textit{(i)} sociolinguistic failures (e.g., code/register selection), \textit{(ii)} pragmatic infelicities (e.g., discourse markers, vocatives, MWEs/proverbs), and \textit{(iii)} semantic failures (e.g., named entities, polysemy, lexical coverage), enabling a more targeted diagnosis of where models break down. The primary limitations of current LLMs in dialectal Arabic–English translation appear to be linguistic rather than merely computational. We therefore provide additional linguistic analysis in Appendix\S\ref{apend:Linguistic Insights}.

\subsection{\texorpdfstring{Severity-weighted LQM Scores}{Severity-weighted LQM Scores}}
Table~\ref{tab:lqm_scores} reports severity-weighted LQM scores (0--100) across models and directions. Overall, \texttt{Gemini-2.5-Pro} achieves the strongest performance, ranking first in $9$ of the $12$ evaluated directions. Performance is generally higher and more stable in DA$\to$EN than in EN$\to$DA, consistent with the additional constraint in EN$\to$DA of maintaining the target dialect’s sociolinguistic identity.

Across directions, EN$\to$UAE yields the highest scores for most models, whereas EN$\to$MOR is the most challenging setting, with most models scoring below $40$ (with the \texttt{Gemini} variants as notable exceptions). At the model level, \texttt{Command-R7B} exhibits the weakest overall performance, with notably low scores on EN$\to$UAE ($21.94$) and EN$\to$MOR ($17.00$). \texttt{Fanar-9B} shows substantial variance across dialects, performing strongly on EN$\to$UAE ($78.32$) but dropping sharply on EN$\to$MOR ($6.46$). Finally, several other models lead in specific directions (e.g., \textit{Command-A} on JOR$\to$EN, \texttt{Gemini-2.5-Flash} on MOR$\to$EN, and \texttt{Fanar-9B} on EN$\to$MAU), suggesting that relative strengths depend on both direction and dialect.

\subsection{Correlations Between Automatic and Human Evaluation}
Table~\ref{tab:spbleu_scores} reports spBLEU scores per direction and model. We compute the correlation between spBLEU and human-derived LQM scores to quantify the agreement between surface-based metrics and expert judgments. We observe a weak positive association (Pearson $r=0.289$, Spearman $\rho=0.322$, $p<0.001$), indicating that spBLEU captures only a limited portion of the variance in severity-weighted human assessments. This gap is expected in our setting, where major quality degradations often stem from sociolinguistic and pragmatic phenomena (e.g., dialect/register mismatches, vocatives/honorifics, discourse markers) that are not well-modeled by n-gram overlap. In Appendix~\ref{sec:lqm_robustness}, we provide additional experiments examining the robustness of LQM to sentence length.

\subsection{Inter-Annotator Agreement (IAA)}
We report IAA using three complementary types of scores: span-detection scores, label-agreement scores on overlapping spans, and chance-corrected agreement measured with Cohen's kappa ($\kappa$). The analysis was conducted on 377 doubly annotated items from both translation directions. Following span-based MT evaluation work~\cite{yeom2025tagged}, we use overlap-based span agreement as the primary detection metric and exact span F1 as a stricter secondary metric. Agreement was strong under overlap-based matching (character-level F1 = 0.760; overlap span F1 = 0.821) but lower under exact span matching (F1 = 0.440), suggesting that annotators usually identified the same error regions while differing in precise span boundaries. On overlapping spans, agreement was highest for coarse error category (F1 = 0.662; $\kappa$ = 0.681), followed by severity (F1 = 0.630; $\kappa$ = 0.427) and category+severity (F1 = 0.517; $\kappa$ = 0.509). Fine-grained error types were more variable (F1 = 0.484), and the strictest criterion---joint agreement on span, error type, and severity---yielded F1 = 0.388. Overall, the results indicate reliable error detection and coarse categorization, with greater variability in fine-grained labeling.




\subsection{Analysis of Error Distributions and Model Attribution}
In Table \ref{Error-distibutions-per-category+M}, we analyze the human-annotated corpus along two axes: \textit{(i)} model-wise contribution to the total error mass, and \textit{(ii)} the distribution of error types by translation direction. Figure~\ref{Fine-grained-LQM} (Appendix) summarizes model-wise error contributions and presents the overall distribution across LQM categories and fine-grained subcategories.

\paragraph{Model-wise error contribution.}
In DA$\to$EN, \texttt{Command-R} and \texttt{Fanar} contribute the largest share of errors, with \texttt{Command-R} peaking in Egyptian ($26.7$\%) and \texttt{Fanar} peaking in Moroccan ($23.3$\%) and Emirati ($23.8$\%). In contrast, \textit{Gemini-2.5-Pro} consistently contributes the smallest error mass across dialects (e.g., $8.5$\% in JOR$\to$ENG). In EN$\to$DA, \texttt{Fanar}, \texttt{Command-R}, and \texttt{Gemma} dominate the error mass. Peaks are observed for \texttt{Command-R} in EN$\to$UAE ($34.4$\%), \texttt{Fanar} in EN$\to$MOR ($33.0$\%), and \texttt{Gemma} in EN$\to$MAU ($22.7$\%). \textit{Gemini-2.5-Pro} continues to contribute substantially less, reaching as low as $2.1$\% of the error mass in EN$\to$MOR and $7.0$\% in EN$\to$EGY.

\paragraph{Error typology by direction.}
We observe a direction-dependent shift in the dominant failure mode. In DA$\to$EN, errors are primarily semantic: semantic categories account for $67.7$\% (Egyptian) up to $92.3$\% (Mauritanian) of the error mass, with particularly high concentrations in Maghrebi dialects (Mauritanian $92.3$\%, Moroccan $85.1$\%). Pragmatic errors form a notable secondary cluster in dialects such as Egyptian ($24.3$\%) and Jordanian ($14.8$\%), indicating difficulties with communicative intent even when the literal meaning is partially recovered. Morphosyntactic and orthographic errors are comparatively rare in DA$\to$EN (often $\leq 5.4$\%), consistent with target-side normalization when generating English.

In EN$\to$DA, the distribution shifts significantly toward sociolinguistic failures. This shift is most pronounced in EN$\to$MAU, where sociolinguistic errors account for $81.5$\% of the error mass. Similar patterns are observed in EN$\to$UAE ($70.6$\%) and EN$\to$PAL ($58.5$\%). However, in Egyptian and Moroccan, semantic errors remain a high secondary failure mode ($39.9$\% and $35.9$\%, respectively). We also observe increases in morphosyntactic and orthographic errors in generation (e.g., morphosyntax up to $19.2$\% for Moroccan; orthography $12.4$\% for Palestinian). Appendix \ref{LQM Fine-Grained} provides a more detailed analysis of the fine-grained error distributions per dialect.

\section{Conclusion}
We presented LQM, a linguistically grounded framework for MT evaluation that organizes errors by linguistic level, enabling diagnostic analysis in diglossic and dialect-rich settings. Crucially, while our empirical evaluation centers on Arabic, the underlying taxonomy is inherently language-agnostic and readily adaptable to other linguistic contexts. In a case study of seven Arabic dialects, we observed a direction-dependent shift in failure modes: DA$\to$EN is largely driven by semantic breakdowns (67.7--92.3\%), reflecting challenges in lexical coverage and meaning transfer from dialectal input, whereas EN$\to$DA is dominated by sociolinguistic failures, with systems often defaulting to dialect-faithful varieties instead of maintaining the target dialect. Improving dialectal MT requires optimizing \textit{both} semantic adequacy and sociolinguistic fidelity.

We also found that spBLEU aligns only weakly with severity-weighted LQM scores (Pearson $r=0.289$, Spearman $\rho=0.322$), consistent with n-gram overlap's insensitivity to pragmatic and dialect-identity errors emphasized by human annotation.
\section*{Acknowledgments}\label{sec:acknow}
We acknowledge support from Canada Research Chairs (CRC), the Natural Sciences and Engineering Research Council of Canada (NSERC; RGPIN-2018-04267), the Social Sciences and Humanities Research Council of Canada (SSHRC; 895-2020-1004), the Canadian Foundation for Innovation (CFI; 37771), the Digital Research Alliance of Canada,\footnote{\href{https://alliancecan.ca}{https://alliancecan.ca}} and UBC ARC-Sockeye.\footnote{\href{https://arc.ubc.ca/ubc-arc-sockeye}{https://arc.ubc.ca/ubc-arc-sockeye}}\\

We thank Aisha Alraeesi for annotating the Emirati Arabic data translated from English and Najla Hassan for annotating the Palestinian Arabic data translated from English. We are also grateful to Yayhay Mohamed Elhaj and Sidi Ebidi for annotating the English-to-Mauritanian direction, to Alcides Alcoba for support with the annotation platform, and to Abderahim Elmadany for his feedback. Finally, we thank the external linguists Ranada Hassan and Saudi Sadiq for independently validating the LQM framework.
\section{Limitations}
Our study has several limitations.

\begin{itemize}
  \item \textbf{Dialect and direction coverage:} Although our study covers seven Arabic varieties overall, we report EN$\to$DA evaluation for only five dialects—Egyptian, Emirati, Mauritanian, Moroccan, and Palestinian. We exclude Jordanian and Yemeni because we were unable to secure enough qualified native-speaker translators to validate outputs in those target dialects. Therefore, our conclusions about dialect preservation in the EN$\to$DA setting apply only to the five evaluated varieties.

    \item \textbf{Domain specificity:} The corpus is derived from transcribed TV dialog, which is conversational and culturally grounded. Although this setting is well-suited for eliciting dialect- and pragmatics-related errors, results may differ in more formal or specialized domains (e.g., legal, medical, or technical translation) with distinct terminology and register constraints.

    \item \textbf{Non-standardized orthography:} Arabic dialect writing lacks fully standardized conventions. We accept spellings judged by native annotators to be common in informal digital contexts, but orthographic variation can complicate consistent span-level localization and make automatic evaluation less straightforward.

    \item \textbf{Prompting and inference settings:} All models are evaluated in a zero-shot setting with a fixed prompt. We do not systematically study the effects of alternative prompting strategies (e.g., few-shot exemplars, constrained output formats) or inference-time controls, which may change both overall quality and the distribution of LQM error types.
\end{itemize}

\section*{Ethical Considerations}
This work studies MT quality for Arabic dialects and introduces LQM, a linguistically grounded framework for span-level error annotation. Because dialectal data can reflect speakers’ regional and social identities, we minimize the risk of sensitive attribute inference by \textit{(i)} reporting results at the dialect/variety level rather than attempting to infer or annotate personal attributes such as gender, age, socioeconomic status, or education, and \textit{(ii)} restricting annotations to translation errors and linguistic phenomena observable in text (e.g., code/register selection, pragmatic appropriateness, semantics, and form-level issues). 

Our corpus is derived from previously released material and contains conversational content; we acknowledge that such data may include culturally specific expressions or potentially sensitive topics. Annotators were instructed to focus on translation quality rather than judge speakers or communities, and to provide brief explanations only when needed for clarity. We also recognize that resources for dialectal MT can be misused to generate targeted or stereotyped content; to mitigate this, we provide documentation emphasizing appropriate use and the limitations of automatic metrics for dialect identity and pragmatics, and we avoid presenting LQM as a tool for profiling individuals.

\bibliography{custom}

@inproceedings{Fernandes2023TheDI,
  title={The Devil Is in the Errors: Leveraging Large Language Models for Fine-grained Machine Translation Evaluation},
  author={Patrick Fernandes and Daniel Deutsch and Mara Finkelstein and Parker Riley and Andr{\'e} F. T. Martins and Graham Neubig and Ankush Garg and J. Clark and Markus Freitag and Orhan Firat},
  booktitle={Conference on Machine Translation},
  year={2023},
  url={https://api.semanticscholar.org/CorpusID:260886800}
}

@inproceedings{Simard2007StatisticalPP,
  title={Statistical Phrase-Based Post-Editing},
  author={Michel Simard and Cyril Goutte and Pierre Isabelle},
  booktitle={North American Chapter of the Association for Computational Linguistics},
  year={2007},
  url={https://api.semanticscholar.org/CorpusID:7695334}
}

@inproceedings{Bhattacharyya2023FindingsOT,
  title={Findings of the WMT 2023 Shared Task on Automatic Post-Editing},
  author={Pushpak Bhattacharyya and Rajen Chatterjee and Markus Freitag and Diptesh Kanojia and Matteo Negri and Marco Turchi},
  booktitle={Conference on Machine Translation},
  year={2023},
  url={https://api.semanticscholar.org/CorpusID:265607923}
}

@inproceedings{Raunak2023LeveragingGF,
  title={Leveraging GPT-4 for Automatic Translation Post-Editing},
  author={Vikas Raunak and Amr Sharaf and Hany Hassan Awadallah and Arul Menezes},
  booktitle={Conference on Empirical Methods in Natural Language Processing},
  year={2023},
  url={https://api.semanticscholar.org/CorpusID:258865299}
}

@article{He2024ImprovingMT,
  title={Improving Machine Translation with Human Feedback: An Exploration of Quality Estimation as a Reward Model},
  author={Zhiwei He and Xing Wang and Wenxiang Jiao and Zhuosheng Zhang and Rui Wang and Shuming Shi and Zhaopeng Tu},
  journal={ArXiv},
  year={2024},
  volume={abs/2401.12873},
  url={https://api.semanticscholar.org/CorpusID:267095196}
}

@inproceedings{jung2024explainable,
  title={EXPLAINABLE CED: A Dataset for Explainable Critical Error Detection in Machine Translation},
  author={Jung, Dahyun and Eo, Sugyeong and Park, Chanjun and Lim, Heui-Seok},
  booktitle={Proceedings of the 2024 Conference of the North American Chapter of the Association for Computational Linguistics: Human Language Technologies (Volume 4: Student Research Workshop)},
  pages={25--35},
  year={2024},
  url={https://aclanthology.org/2024.naacl-srw.4.pdf}
}

@inproceedings{wasti2025translationcorrect,
  title={TranslationCorrect: A Human-Centered Post-Editing Framework for Error-Aware Machine Translation},
  author={Wasti, Adnan and Lee, Matthew and Alam, Tausifa and Ghosh, Sreyasi and Carpuat, Marine},
  booktitle={Proceedings of ACL 2025 (to appear)},
  year={2025},
  url={https://arxiv.org/abs/2506.18337}
}

@article{yakhni2025lebanese,
  title={Fine-tuning Arabic LLMs for Lebanese Dialect Translation and Evaluation},
  author={Yakhni, Malak and Chehab, Jeanine},
  journal={arXiv preprint arXiv:2405.12534},
  year={2025}, 
  url={https://arxiv.org/pdf/2505.00114?}
}

@inproceedings{kadaoui-etal-2023-tarjamat,
    title = "{TARJAMAT}: Evaluation of Bard and {C}hat{GPT} on Machine Translation of Ten {A}rabic Varieties",
    author = "Kadaoui, Karima  and
      Magdy, Samar M.  and
      Waheed, Abdul  and
      Khondaker, Md Tawkat Islam  and
      El-Shangiti, Ahmed Oumar  and
      Nagoudi, El Moatez Billah  and
      Abdul-Mageed, Muhammad",
    editor = "Sawaf, Hassan  and
      El-Beltagy, Samhaa  and
      Zaghouani, Wajdi  and
      Magdy, Walid  and
      Abdelali, Ahmed  and
      Tomeh, Nadi  and
      Abu Farha, Ibrahim  and
      Habash, Nizar  and
      Khalifa, Salam  and
      Keleg, Amr  and
      Haddad, Hatem  and
      Zitouni, Imed  and
      Mrini, Khalil  and
      Almatham, Rawan",
    booktitle = "Proceedings of ArabicNLP 2023",
    month = dec,
    year = "2023",
    address = "Singapore (Hybrid)",
    publisher = "Association for Computational Linguistics",
    url = "https://aclanthology.org/2023.arabicnlp-1.6/",
    doi = "10.18653/v1/2023.arabicnlp-1.6",
    pages = "52--75"
}

@article{alves2024xtower,
  title={xTOWER: Multilingual Translation Error Explanation and Correction with Large Language Models},
  author={Alves, Duarte M and Pombal, José P and Guerreiro, Nuno M and Martins, André FT},
  journal={arXiv preprint arXiv:2406.19482},
  year={2024},
  url={https://aclanthology.org/2024.findings-emnlp.892.pdf}
}

@inproceedings{bouamor2018madar,
  title={The MADAR Arabic dialect corpus and lexicon},
  author={Bouamor, Houda and Habash, Nizar and Salameh, Mohammad and Zaghouani, Wajdi and Rambow, Owen and Abdulrahim, Dana and Obeid, Ossama and Khalifa, Salam and Eryani, Fadhl and Erdmann, Alexander and others},
  booktitle={Proceedings of the eleventh international conference on language resources and evaluation (LREC 2018)},
  year={2018},
  url={https://aclanthology.org/L18-1535.pdf}
}

@inproceedings{habash2012conventional,
  title={Conventional orthography for dialectal Arabic.},
  author={Habash, Nizar and Diab, Mona T and Rambow, Owen},
  booktitle={LREC},
  pages={711--718},
  year={2012},
  url={http://lrec.elra.info/proceedings/lrec2012/pdf/579_Paper.pdf}
}

@inproceedings{zbib2012machine,
  title={Machine Translation of Arabic Dialects},
  author={Zbib, Rabih and Matsoukas, Spyros and Schwartz, Richard and Makhoul, John and Jimenez, Enrique and Malarkey, Chad},
  booktitle={Proceedings of the 2012 Conference of the North American Chapter of the Association for Computational Linguistics (NAACL)},
  year={2012},
  url={https://aclanthology.org/N12-1006.pdf}
}

@inproceedings{sajjad-etal-2020-arabench,
    title = "{A}ra{B}ench: Benchmarking Dialectal {A}rabic-{E}nglish Machine Translation",
    author = "Sajjad, Hassan  and
      Abdelali, Ahmed  and
      Durrani, Nadir  and
      Dalvi, Fahim",
    editor = "Scott, Donia  and
      Bel, Nuria  and
      Zong, Chengqing",
    booktitle = "Proceedings of the 28th International Conference on Computational Linguistics",
    month = dec,
    year = "2020",
    address = "Barcelona, Spain (Online)",
    publisher = "International Committee on Computational Linguistics",
    url = "https://aclanthology.org/2020.coling-main.447/",
    doi = "10.18653/v1/2020.coling-main.447",
    pages = "5094--5107"
}

@inproceedings{Lommel2014MultidimensionalQM,
  title={Multidimensional Quality Metrics (MQM): A Framework for Declaring and Describing Translation Quality Metrics},
  author={Arle Lommel and Hans Uszkoreit and Aljoscha Burchardt},
    booktitle={Proceedings of the Ninth International Conference on Language Resources and Evaluation (LREC'14)},
  year={2014},
  url={https://api.semanticscholar.org/CorpusID:55606096}
}

@article{Liu2025NewTF,
  title={New Trends for Modern Machine Translation with Large Reasoning Models},
  author={Sinuo Liu and Chenyang Lyu and Minghao Wu and Longyue Wang and Weihua Luo and Kaifu Zhang and Zifu Shang},
  journal={ArXiv},
  year={2025},
  volume={abs/2503.10351},
  url={https://api.semanticscholar.org/CorpusID:276961699}
}

@inproceedings{wang2025drt,
  title={Drt: Deep reasoning translation via long chain-of-thought},
  author={Wang, Jiaan and Meng, Fandong and Liang, Yunlong and Zhou, Jie},
  booktitle={Findings of the Association for Computational Linguistics: ACL 2025},
  pages={6770--6782},
  year={2025}, 
  url={https://aclanthology.org/2025.findings-acl.351.pdf}
}

@article{He2025R1T1FI,
  title={R1-T1: Fully Incentivizing Translation Capability in LLMs via Reasoning Learning},
  author={Minggui He and Yilun Liu and Shimin Tao and Yuanchang Luo and Hongyong Zeng and Chang Su and Li Zhang and Hongxia Ma and Daimeng Wei and Weibin Meng and Hao Yang and Boxing Chen and Osamu Yoshie},
  journal={ArXiv},
  year={2025},
  volume={abs/2502.19735},
  url={https://api.semanticscholar.org/CorpusID:276647922}
}

@article{wang2025deep,
  title={Deep reasoning translation via reinforcement learning},
  author={Wang, Jiaan and Meng, Fandong and Zhou, Jie},
  journal={arXiv preprint arXiv:2504.10187},
  year={2025},
  url={https://arxiv.org/pdf/2504.10187}
}

@article{Feng2025MTR1ZeroAL,
  title={MT-R1-Zero: Advancing LLM-based Machine Translation via R1-Zero-like Reinforcement Learning},
  author={Zhaopeng Feng and Shaosheng Cao and Jiahan Ren and Jiayuan Su and Ruizhe Chen and Yan Zhang and Zhe Xu and Yao Hu and Jian Wu and Zuozhu Liu},
  journal={ArXiv},
  year={2025},
  volume={abs/2504.10160},
  url={https://api.semanticscholar.org/CorpusID:277780438}
}

@article{Kocmi2024ErrorSA,
  title={Error Span Annotation: A Balanced Approach for Human Evaluation of Machine Translation},
  author={Tom Kocmi and Vil{\'e}m Zouhar and Eleftherios Avramidis and Roman Grundkiewicz and Marzena Karpinska and Maja Popovi'c and Mrinmaya Sachan and Mariya Shmatova},
  journal={ArXiv},
  year={2024},
  volume={abs/2406.11580},
  url={https://api.semanticscholar.org/CorpusID:270559821}
}

@article{Freitag2021ExpertsEA,
  title={Experts, Errors, and Context: A Large-Scale Study of Human Evaluation for Machine Translation},
  author={Markus Freitag and George F. Foster and David Grangier and Viresh Ratnakar and Qijun Tan and Wolfgang Macherey},
  journal={Transactions of the Association for Computational Linguistics},
  year={2021},
  volume={9},
  pages={1460-1474},
  url={https://api.semanticscholar.org/CorpusID:233444275}
}

@inproceedings{Kreutzer2020CorrectMI,
  title={Correct Me If You Can: Learning from Error Corrections and Markings},
  author={Julia Kreutzer and Nathaniel Berger and Stefan Riezler},
  booktitle={European Association for Machine Translation Conferences/Workshops},
  year={2020},
  url={https://api.semanticscholar.org/CorpusID:216080598}
}

@article{team2024gemma,
  title={Gemma 2: Improving open language models at a practical size},
  author={Team, Gemma and Riviere, Morgane and Pathak, Shreya and Sessa, Pier Giuseppe and Hardin, Cassidy and Bhupatiraju, Surya and Hussenot, L{\'e}onard and Mesnard, Thomas and Shahriari, Bobak and Ram{\'e}, Alexandre and others},
  journal={arXiv preprint arXiv:2408.00118},
  year={2024},
  url={https://arxiv.org/pdf/2408.00118?}
}

@article{team2025fanar,
  title={Fanar: An arabic-centric multimodal generative ai platform},
  author={Team, Fanar and Abbas, Ummar and Ahmad, Mohammad Shahmeer and Alam, Firoj and Altinisik, Enes and Asgari, Ehsannedin and Boshmaf, Yazan and Boughorbel, Sabri and Chawla, Sanjay and Chowdhury, Shammur and others},
  journal={arXiv preprint arXiv:2501.13944},
  year={2025}, 
  url={https://arxiv.org/pdf/2501.13944?}
}

@article{team2023gemini,
  title={Gemini: a family of highly capable multimodal models},
  author={Team, Gemini and Anil, Rohan and Borgeaud, Sebastian and Alayrac, Jean-Baptiste and Yu, Jiahui and Soricut, Radu and Schalkwyk, Johan and Dai, Andrew M and Hauth, Anja and Millican, Katie and others},
  journal={arXiv preprint arXiv:2312.11805},
  year={2023}, 
  url={https://arxiv.org/pdf/2312.11805}
}

@software{cohere2024commandrplus,
  author       = {CohereLabs},
  title        = {{Command R+}: A 104B Parameter Open-Weight Model for RAG and Tool Use},
  month        = apr,
  year         = {2024},
  publisher    = {Hugging Face},
  url          = {https://huggingface.co/CohereLabs/c4ai-command-r-plus}
}

@inproceedings{lu2025mqm,
  title={MQM-APE: toward high-quality error annotation predictors with automatic post-editing in LLM translation evaluators},
  author={Lu, Qingyu and Ding, Liang and Zhang, Kanjian and Zhang, Jinxia and Tao, Dacheng},
  booktitle={Proceedings of the 31st International Conference on Computational Linguistics},
  pages={5570--5587},
  year={2025},   
  url={https://aclanthology.org/2025.coling-main.374.pdf}
}

@article{Kocmi2023GEMBAMQMDT,
  title={GEMBA-MQM: Detecting Translation Quality Error Spans with GPT-4},
  author={Tom Kocmi and Christian Federmann},
  journal={ArXiv},
  year={2023},
  volume={abs/2310.13988},
  url={https://api.semanticscholar.org/CorpusID:264426679}
}

@article{Guerreiro2023xcometTM,
  title={xcomet: Transparent Machine Translation Evaluation through Fine-grained Error Detection},
  author={Nuno M. Guerreiro and Ricardo Rei and Daan van Stigt and Lu{\'i}sa Coheur and Pierre Colombo and Andr{\'e} Martins},
  journal={Transactions of the Association for Computational Linguistics},
  year={2023},
  volume={12},
  pages={979-995},
  url={https://api.semanticscholar.org/CorpusID:264146484}
}

@article{Zhan2025AreLR,
  title={Are Large Reasoning Models Good Translation Evaluators? Analysis and Performance Boost},
  author={Runzhe Zhan and Zhihong Huang and Xinyi Yang and Lidia S. Chao and Min Yang and Derek F. Wong},
  journal={ArXiv},
  year={2025},
  volume={abs/2510.20780},
  url={https://api.semanticscholar.org/CorpusID:282304185}
}

@inproceedings{talafha-etal-2024-casablanca,
    title = "{C}asablanca: Data and Models for Multidialectal {A}rabic Speech Recognition",
    author = "Talafha, Bashar  and
      Kadaoui, Karima  and
      Magdy, Samar Mohamed  and
      Habiboullah, Mariem  and
      Chafei, Chafei Mohamed  and
      El-Shangiti, Ahmed Oumar  and
      Zayed, Hiba  and
      Tourad, Mohamedou Cheikh  and
      Alhamouri, Rahaf  and
      Assi, Rwaa  and
      Alraeesi, Aisha  and
      Mohamed, Hour  and
      Alwajih, Fakhraddin  and
      Mohamed, Abdelrahman  and
      El Mekki, Abdellah  and
      Nagoudi, El Moatez Billah  and
      Saadia, Benelhadj Djelloul Mama  and
      Alsayadi, Hamzah A.  and
      Al-Dhabyani, Walid  and
      Shatnawi, Sara  and
      Ech-chammakhy, Yasir  and
      Makouar, Amal  and
      Berrachedi, Yousra  and
      Jarrar, Mustafa  and
      Shehata, Shady  and
      Berrada, Ismail  and
      Abdul-Mageed, Muhammad",
    editor = "Al-Onaizan, Yaser  and
      Bansal, Mohit  and
      Chen, Yun-Nung",
    booktitle = "Proceedings of the 2024 Conference on Empirical Methods in Natural Language Processing",
    month = nov,
    year = "2024",
    address = "Miami, Florida, USA",
    publisher = "Association for Computational Linguistics",
    url = "https://aclanthology.org/2024.emnlp-main.1211/",
    doi = "10.18653/v1/2024.emnlp-main.1211",
    pages = "21745--21758"
}

@inproceedings{abdul-mageed-etal-2024-nadi,
    title = "{NADI} 2024: The Fifth Nuanced {A}rabic Dialect Identification Shared Task",
    author = "Abdul-Mageed, Muhammad  and
      Keleg, Amr  and
      Elmadany, AbdelRahim  and
      Zhang, Chiyu  and
      Hamed, Injy  and
      Magdy, Walid  and
      Bouamor, Houda  and
      Habash, Nizar",
    editor = "Habash, Nizar  and
      Bouamor, Houda  and
      Eskander, Ramy  and
      Tomeh, Nadi  and
      Abu Farha, Ibrahim  and
      Abdelali, Ahmed  and
      Touileb, Samia  and
      Hamed, Injy  and
      Onaizan, Yaser  and
      Alhafni, Bashar  and
      Antoun, Wissam  and
      Khalifa, Salam  and
      Haddad, Hatem  and
      Zitouni, Imed  and
      AlKhamissi, Badr  and
      Almatham, Rawan  and
      Mrini, Khalil",
    booktitle = "Proceedings of the Second Arabic Natural Language Processing Conference",
    month = aug,
    year = "2024",
    address = "Bangkok, Thailand",
    publisher = "Association for Computational Linguistics",
    url = "https://aclanthology.org/2024.arabicnlp-1.79/",
    doi = "10.18653/v1/2024.arabicnlp-1.79",
    pages = "709--728"
}

@article{Yeom2025TaggedSA,
  title={Tagged Span Annotation for Detecting Translation Errors in Reasoning LLMs},
  author={Taemin Yeom and Yonghyun Ryu and Yoonjung Choi and Jinyeong Bak},
  journal={Proceedings of the Tenth Conference on Machine Translation},
  year={2025},
  url={https://api.semanticscholar.org/CorpusID:282902140}
}

@inproceedings{popovic-2017-chrf,
    title = "chr{F}++: words helping character n-grams",
    author = "Popovi{\'c}, Maja",
    editor = "Bojar, Ond{\v{r}}ej  and
      Buck, Christian  and
      Chatterjee, Rajen  and
      Federmann, Christian  and
      Graham, Yvette  and
      Haddow, Barry  and
      Huck, Matthias  and
      Yepes, Antonio Jimeno  and
      Koehn, Philipp  and
      Kreutzer, Julia",
    booktitle = "Proceedings of the Second Conference on Machine Translation",
    month = sep,
    year = "2017",
    address = "Copenhagen, Denmark",
    publisher = "Association for Computational Linguistics",
    url = "https://aclanthology.org/W17-4770/",
    doi = "10.18653/v1/W17-4770",
    pages = "612--618"
}

@inproceedings{10.3115/1073083.1073135,
author = {Papineni, Kishore and Roukos, Salim and Ward, Todd and Zhu, Wei-Jing},
title = {BLEU: a method for automatic evaluation of machine translation},
year = {2002},
publisher = {Association for Computational Linguistics},
address = {USA},
url = {https://doi.org/10.3115/1073083.1073135},
doi = {10.3115/1073083.1073135},
booktitle = {Proceedings of the 40th Annual Meeting on Association for Computational Linguistics},
pages = {311–318},
numpages = {8},
location = {Philadelphia, Pennsylvania},
series = {ACL '02}
}

@article{goyal-etal-2022-flores,
    title = "The {F}lores-101 Evaluation Benchmark for Low-Resource and Multilingual Machine Translation",
    author = "Goyal, Naman  and
      Gao, Cynthia  and
      Chaudhary, Vishrav  and
      Chen, Peng-Jen  and
      Wenzek, Guillaume  and
      Ju, Da  and
      Krishnan, Sanjana  and
      Ranzato, Marc{'}Aurelio  and
      Guzm{\'a}n, Francisco  and
      Fan, Angela",
    editor = "Roark, Brian  and
      Nenkova, Ani",
    journal = "Transactions of the Association for Computational Linguistics",
    volume = "10",
    year = "2022",
    address = "Cambridge, MA",
    publisher = "MIT Press",
    url = "https://aclanthology.org/2022.tacl-1.30/",
    doi = "10.1162/tacl_a_00474",
    pages = "522--538"}

@inproceedings{rei-etal-2020-comet,
    title = "{COMET}: A Neural Framework for {MT} Evaluation",
    author = "Rei, Ricardo  and
      Stewart, Craig  and
      Farinha, Ana C  and
      Lavie, Alon",
    editor = "Webber, Bonnie  and
      Cohn, Trevor  and
      He, Yulan  and
      Liu, Yang",
    booktitle = "Proceedings of the 2020 Conference on Empirical Methods in Natural Language Processing (EMNLP)",
    month = nov,
    year = "2020",
    address = "Online",
    publisher = "Association for Computational Linguistics",
    url = "https://aclanthology.org/2020.emnlp-main.213/",
    doi = "10.18653/v1/2020.emnlp-main.213",
    pages = "2685--2702"}

@article{hymes1972communicative,
  title={On communicative competence. Sociolinguistics},
  author={Hymes, Dell and Pride, JB and Holmes, Janet},
  journal={Eds. Pride, JB y J. Holmes},
  pages={269--293},
  year={1972},
  url={https://d1wqtxts1xzle7.cloudfront.net/33165140/communicative_compentence-libre.pdf?1394281094=&response-content-disposition=inline%3B+filename%3DCommunicative_competence_and_language.pdf&Expires=1768415377&Signature=EkqGLMTele8j-jEU676ib97WhUSnVD2ZyoBOuqK9ZiJa-RILJmWcY8DxH8qZ85FGaULMeXsfQJAJ-Kxdsw65CTD-vvGvTgkhp99b~2f6P6vICc9Wo6dlNLrU864zldaZQNaWRNbppf9xpGtzwfIZG8gRJK1iDvzsygmjpceRPHyGc0zdqPoUjWis65EDE9A4mzxgDO4X6NYeuF9uZPDmgkqUJ~sozmPsoi3Lhv4hMM6ggJw-4TiR7PAT5xyeoZkRUBugQ5q6lg62WEgrrF~81moIHzVQOc3xQtCsXtkLowsEe9vewEvRep90mMdqEUS3nTkz5E2IOTYqq2Mhnu5Wfw__&Key-Pair-Id=APKAJLOHF5GGSLRBV4ZA}
}

@book{halliday1978language,
  title={Language as social semiotic},
  author={Halliday, Michael AK},
  year={1978},
  publisher={London Arnold},
  url={https://www.torrossa.com/en/resources/an/5015200#page=274}
}

@book{austin1975things,
  title={How to do things with words},
  author={Austin, John Langshaw},
  year={1975},
  publisher={Harvard university press},
  url={https://pure.mpg.de/rest/items/item_2271128_6/component/file_2271430/content}
}

@book{searle1969speech,
  title={Speech acts: An essay in the philosophy of language},
  author={Searle, John R},
  year={1969},
  publisher={Cambridge university press},
  url={https://books.google.ca/books?hl=en&lr=&id=4UKbAAAAQBAJ&oi=fnd&pg=PA1&dq=Speech+acts:+An+essay+in+the+philosophy+of+language&ots=u5wr5zOzv0&sig=Mtns6eTPVdYtzu-KxHxLm-wSClY#v=onepage&q=Speech%20acts%3A%20An%20essay%20in%20the%20philosophy%20of%20language&f=false}
}

@article{nunberg1994idioms,
  title={Idioms},
  author={Nunberg, Geoffrey and Sag, Ivan A and Wasow, Thomas},
  journal={Language},
  volume={70},
  number={3},
  pages={491--538},
  year={1994},
  publisher={Linguistic Society of America}, 
  url={https://muse.jhu.edu/pub/24/article/452933/summary}
}

@article{fraser1999discourse,
  title={What are discourse markers?},
  author={Fraser, Bruce},
  journal={Journal of pragmatics},
  volume={31},
  number={7},
  pages={931--952},
  year={1999},
  publisher={Elsevier},
  url={https://pdf.sciencedirectassets.com/271806/1-s2.0-S0378216600X00549/1-s2.0-S0378216698001015/main.pdf?X-Amz-Security-Token=IQoJb3JpZ2luX2VjEEkaCXVzLWVhc3QtMSJHMEUCIG2sEvRSpOPMJ25Fpbq51Dyd%2FECDYrezpcHGmNbrj2jzAiEAhq1lmTpplsMOAyevqIhox5Lx6GhaEvlBnUFN1wNErPgqswUIERAFGgwwNTkwMDM1NDY4NjUiDJc6huWOsQgfWOYeayqQBZSmXra1Fe56nv5oHejyYeN6ImQkqFy64S6jkUP99ILt%2FvarVTWIe2%2BaPT5gAS5LXCBPn3WcpdL%2Frh%2BfxUP%2BbmetucXSAVK8B4whTyiSBNJ4DFKqENmk4NDiLRM0V8NEwDhKsT%2BOoN0aSY5%2FHA6XBqgmDcTaM1KwuqW%2Ft9zJg4prRNETL0uDyZy6CqHpLRbMsbkrLdI3Ja7lfyO2I8XHuRnTZYOqWpnXO1L2b1q8YocUEbPAnLCBuWYjhKyVft%2BRbukr%2FfVe6ors6Nxv2vcfciwjXPnbSAbLd35nkBPvGgCnL6SuxUsM2zNzWIUWCSi27Y7IFLPQrN9xgEs6LsOUvhPHoUBz4D%2BNiKZgrp8zSx%2B1UbmjhMxz1JgiGsSc93rNFRv2VrUkeHNbSu923D3U6kGLTAgtMCbXJ1yCzNwa2gmRgpWxQYkOOy4ymc1g8UzvP8wbumC7PHwM%2FeMW%2FhXRgLJwF2POOIEQTgEgfYCakmPLR9kMGWvyHYIKmjKq5tWgzxnbwFEERKWohPmK92PVnwQEKYO2T73OLUWEv%2BUmKAk3RvY4qLGNwG%2BoikqYxYk%2FegaddR8FhIPEfmuJH3EmBYtxa1PgPNA882Y5J0dSr0hBCZF5k6CSrR4Qid0QDCg6hhRgMNMGbfKY14hQ9TKtIW0xNruknVY4K%2BuSfTTi96vSn2xMsungzdrSeg6H%2Fo5UrRIfV0zEzRM5r7q4ZfMw5cmMc1L2UHv4asCzP2acDNhqsg0HfJiZtyusxloJvSJA3L8j6JHyr1p%2FP1aeqoGZMIyCMCjc6%2FB1mZwwremVmdck9tN%2Bjr%2Bd4jiKa9D2Ib2ZZv7MVTPBU0C7AS%2BLRw98gFL90ekF1UGoHNGHUycNTF2UMN3Cm8sGOrEB%2BJoPvUMOSiy%2BAHpzau%2FSJ2R9HkUK4j2p1oeuwAT6owZPnk7nGfAw2XKHRUNKNISY%2BPu1qLCQeGbJO9a56a38jFh%2BrOLgTB5p6%2BqVsU6F0kkEDQHYF%2BVpg3qC7vQc91vGLK8L2%2ByFA8jNvhSYAXtYvYbcIjso5fuIBROwKH6mYDhjdpuCayYzLBinye9Xo3hzITSwXiDTdC9at4Qig0DTm2DK1aPj9G730kkpBB8UMCit&X-Amz-Algorithm=AWS4-HMAC-SHA256&X-Amz-Date=20260114T012326Z&X-Amz-SignedHeaders=host&X-Amz-Expires=300&X-Amz-Credential=ASIAQ3PHCVTY34LFETOM%2F20260114%2Fus-east-1%2Fs3%2Faws4_request&X-Amz-Signature=a713082f2c085bb99e714f402f0c0ca588aed54d5cb026128f4c2bbddcbd9e8e&hash=800f736a88bc10ae62abfc6dec16922a444b682a28163ef6db2c2f8261e457aa&host=68042c943591013ac2b2430a89b270f6af2c76d8dfd086a07176afe7c76c2c61&pii=S0378216698001015&tid=spdf-24633ba7-e4a0-49aa-95ec-f679b85ecd2b&sid=b25a85c77a9844454589a467e840eae019d4gxrqa&type=client&tsoh=d3d3LnNjaWVuY2VkaXJlY3QuY29t&rh=d3d3LnNjaWVuY2VkaXJlY3QuY29t&ua=19075600035f5b56090e51&rr=9bd953fa3d0d8425&cc=ca}
}

@article{farwell1999pragmatics,
  title={Pragmatics and translation},
  author={Farwell, David and Helmreich, Stephen},
  journal={Procesamiento del lenguaje natural, n{\textordmasculine} 24 (mayo 1999); pp. 18-39},
  year={1999},
  publisher={Sociedad Espa{\~n}ola para el Procesamiento del Lenguaje Natural},
  url={https://rua.ua.es/server/api/core/bitstreams/e8b7e39a-4554-4247-87b7-dcc67c4db4c4/content}
}

@book{cruse1986lexical,
  title={Lexical semantics},
  author={Cruse, D Alan},
  year={1986},
  publisher={Cambridge university press},
  url={https://books.google.ca/books?hl=en&lr=&id=xDSBaet2uSsC&oi=fnd&pg=PR11&dq=Lexical+semantics&ots=9E4dogHNBa&sig=wp3kgM64PpmAkqoOXWIUAikVGYQ#v=onepage&q=Lexical%20semantics&f=false}
}

@article{soames1987direct,
  title={Direct reference, propositional attitudes, and semantic content},
  author={Soames, Scott},
  journal={Philosophical topics},
  volume={15},
  number={1},
  pages={47--87},
  year={1987},
  publisher={JSTOR}, 
  url={https://www.jstor.org/stable/43153993}
}

@book{radford2004english,
  title={English syntax: An introduction},
  author={Radford, Andrew},
  year={2004},
  publisher={Cambridge University Press},
  url={https://books.google.ca/books?hl=en&lr=&id=LdAi292Q4-0C&oi=fnd&pg=PR9&dq=English+syntax:+An+introduction&ots=HGYy4Vt3Jw&sig=obLZCLyOTNI1W4_1uWYtNZyXQaQ#v=onepage&q=English%20syntax%3A%20An%20introduction&f=false}
}

@book{brown2013canonical,
  title={Canonical morphology and syntax},
  author={Brown, Dunstan and Chumakina, Marina and Corbett, Greville G},
  year={2013},
  publisher={Oxford University Press},
  url={https://www.researchgate.net/profile/Dunstan-Brown/publication/265236325_Canonical_Morphology_and_Syntax/links/561ebb6e08aecade1acd0552/Canonical-Morphology-and-Syntax.pdf}
}

@incollection{corbett2006introduction,
  title={Introduction: Canonical agreement},
  author={Corbett, Greville G},
  booktitle={Agreement},
  pages={1--34},
  year={2006},
  publisher={Cambridge University Press},
  url={https://openresearch.surrey.ac.uk/esploro/fulltext/bookChapter/Introduction-Canonical-agreement/99516821502346?repId=12140113850002346&mId=13140353500002346&institution=44SUR_INST}
}

@article{greenberg1963some,
  title={Some universals of grammar with particular reference to the order of meaningful elements},
  author={Greenberg, Joseph H and others},
  journal={Universals of language},
  volume={2},
  pages={73--113},
  year={1963},
  url={http://www.fb10.uni-bremen.de/homepages/hackmack/synsem/pdf/Universals_of_Language.pdf}
}

@book{palmer2001mood,
  title={Mood and modality},
  author={Palmer, Frank Robert},
  year={2001},
  publisher={Cambridge university press}, 
  url={https://books.google.ca/books?hl=en&lr=&id=xKUvDFTARR8C&oi=fnd&pg=PR15&dq=Palmer,+F.+R.+(2001).+Mood+and+Modality.&ots=ml9ILnmnuW&sig=eoPzhNHEgvjINAkJlb5fWHBeKQ8#v=onepage&q=Palmer%2C%20F.%20R.%20(2001).%20Mood%20and%20Modality.&f=false}
}

@article{derwing1992orthographic,
  title={Orthographic aspects of linguistic competence},
  author={Derwing, Bruce L},
  journal={The linguistics of literacy},
  volume={21},
  pages={193--211},
  year={1992},
  url={https://www.torrossa.com/en/resources/an/5015302#page=214}
}

@book{brown1987politeness,
  title={Politeness: Some universals in language usage},
  author={Brown, Penelope and Levinson, Stephen C},
  volume={4},
  year={1987},
  publisher={Cambridge university press},
  url={https://pure.mpg.de/rest/items/item_64421/component/file_2225570/content}
}

@book{mccready2019semantics,
  title={The semantics and pragmatics of honorification: Register and social meaning},
  author={McCready, Elin},
  volume={11},
  year={2019},
  publisher={Oxford University Press},
  url={https://books.google.ca/books?hl=en&lr=&id=x0KjDwAAQBAJ&oi=fnd&pg=PP1&dq=The+semantics+and+pragmatics+of+honorification:+Register+and+social+meaning&ots=495h3kwewa&sig=F1XjCC8-E_IWIocbGXgciew_hPc#v=onepage&q=The%20semantics%20and%20pragmatics%20of%20honorification%3A%20Register%20and%20social%20meaning&f=false}
}

@article{ferguson1959diglossia,
  title={Diglossia},
  author={Ferguson, Charles A},
  journal={word},
  volume={15},
  number={2},
  pages={325--340},
  year={1959},
  publisher={Taylor \& Francis},
  url={https://www.tandfonline.com/doi/pdf/10.1080/00437956.1959.11659702}
}

@book{schiffrin1987discourse,
  title={Discourse markers},
  author={Schiffrin, Deborah},
  number={5},
  year={1987},
  publisher={Cambridge University Press},
  url={https://books.google.ca/books?hl=en&lr=&id=hs7J-WqPtPAC&oi=fnd&pg=PP10&dq=Discourse+markers&ots=HYfIhDTil5&sig=M2-1ScaPSui4AEXn1t-CRuS_dB4#v=onepage&q=Discourse%20markers&f=false}
}

@article{fraser2009account,
  title={An account of discourse markers},
  author={Fraser, Bruce},
  journal={International review of Pragmatics},
  volume={1},
  number={2},
  pages={293--320},
  year={2009},
  publisher={Brill},
  url={https://brill.com/view/journals/irp/1/2/article-p293_3.xml}
}

@book{gumperz1982discourse,
  title={Discourse strategies},
  author={Gumperz, John J},
  number={1},
  year={1982},
  publisher={Cambridge University Press},
  url={https://d1wqtxts1xzle7.cloudfront.net/30244605/81020627-libre.pdf?1363452275=&response-content-disposition=inline%3B+filename%3DDiscourse_strategies.pdf&Expires=1768414958&Signature=dAlUqVFuuHN6T8ecVedjPhi7vMuK93xAjSqKRZ1is2XDg-jRXUCXHVdy7BtTgVGmmxdj9kCkNoaSR3-9kCLsNiIVWk-CmnziFAnHEx5Tn9PuYoFFLmLmo5CZbTW2ALEBOSSCmPoEBC~BMGuDg45sFDszc3uqMmte2HGRYdqBEeKD23A3G0y-ESZwROyKlcyf2bqgMFRnYriiZSHsbFL055KnVaBv9mZuLrmP94wbia3oezQeVG-5zXFSkUiXcpeexw8sDhamO7stzjDafgBBXVPQzqm~51jmqOXZR9PJIRz6yeHxxkh1hHk6b2WabU1rI9kk~EYfMkUCbEZxdTvPHw__&Key-Pair-Id=APKAJLOHF5GGSLRBV4ZA}
}

@book{bassiouney2020arabic,
  title={Arabic sociolinguistics: Topics in diglossia, gender, identity, and politics},
  author={Bassiouney, Reem},
  year={2020},
  publisher={Georgetown University Press},
  url={https://edinburghuniversitypress.com/pub/media/resources/9781474457361_Arabic_Sociolinguistics_-_Chapter_1.pdf}
}

@book{halliday2014cohesion,
  title={Cohesion in english},
  author={Halliday, Michael Alexander Kirkwood and Hasan, Ruqaiya},
  year={2014},
  publisher={Routledge},
  url={https://en.ulis.vnu.edu.vn/files/uploads/2020/07/35-4.pdf#page=185}
}

@book{kamp2013discourse,
  title={From discourse to logic: Introduction to modeltheoretic semantics of natural language, formal logic and discourse representation theory},
  author={Kamp, Hans and Reyle, Uwe},
  volume={42},
  year={2013},
  publisher={Springer Science \& Business Media},
  url={https://books.google.ca/books?hl=en&lr=&id=xYPrCAAAQBAJ&oi=fnd&pg=PA1&dq=From+discourse+to+logic:+Introduction+to+model+theoretic+semantics+of+natural+language,+formal+logic+and+discourse+representation+theory&ots=W3KxbG4gYM&sig=haXm_Fdw4UPXrKz7UspoDbgyVmo#v=onepage&q=From%20discourse%20to%20logic%3A%20Introduction%20to%20model%20theoretic%20semantics%20of%20natural%20language%2C%20formal%20logic%20and%20discourse%20representation%20theory&f=false}
}

@book{levinson1983pragmatics,
  title={Pragmatics},
  author={Levinson, Stephen C},
  year={1983},
  publisher={Cambridge university press}
}

@INPROCEEDINGS{10638285,
  author={Abdul-Nabi, Razan and Obeidat, Rasha and Bsoul, Anas},
  booktitle={2024 15th International Conference on Information and Communication Systems (ICICS)}, 
  title={A Survey on Machine Translation of Low-Resource Arabic Dialects}, 
  year={2024},
  volume={},
  number={},
  pages={1-6},
  doi={10.1109/ICICS63486.2024.10638285}}

@article{mekki2026alexandria,
  title={Alexandria: A Multi-Domain Dialectal Arabic Machine Translation Dataset for Culturally Inclusive and Linguistically Diverse LLMs},
  author={Mekki, Abdellah El and Magdy, Samar M and Atou, Houdaifa and AbuHweidi, Ruwa and Qawasmeh, Baraah and Nacar, Omer and Al-hibiri, Thikra and Saadie, Razan and Alsayadi, Hamzah and Hammouda, Nadia Ghezaiel and others},
  journal={arXiv preprint arXiv:2601.13099},
  year={2026},
  url={https://arxiv.org/abs/2601.13099}
}

@inproceedings{yeom2025tagged,
  title={Tagged Span Annotation for Detecting Translation Errors in Reasoning LLMs},
  author={Yeom, Taemin and Ryu, Yonghyun and Choi, Yoonjung and Bak, JinYeong},
  booktitle={Proceedings of the Tenth Conference on Machine Translation},
  pages={878--886},
  year={2025},
  url={https://aclanthology.org/2025.wmt-1.62.pdf}
}

@article{brewster2025evaluating,
  title={Evaluating human-in-the-loop strategies for artificial intelligence-enabled translation of patient discharge instructions: a multidisciplinary analysis},
  author={Brewster, Ryan CL and Tse, Gabriel and Fan, Angela L and Elborki, Marwa and Newell, Maiah and Gonzalez, Priscilla and Hoq, Amitra and Chang, Crystal and Chowdhury, Maksud and Geeti, Adiba and others},
  journal={NPJ digital medicine},
  volume={8},
  number={1},
  pages={629},
  year={2025},
  publisher={Nature Publishing Group UK London},
  url={https://www.nature.com/articles/s41746-025-02055-6}
}

@inproceedings{yao-etal-2024-benchmarking,
    title = "Benchmarking Machine Translation with Cultural Awareness",
    author = "Yao, Binwei  and
      Jiang, Ming  and
      Bobinac, Tara  and
      Yang, Diyi  and
      Hu, Junjie",
    editor = "Al-Onaizan, Yaser  and
      Bansal, Mohit  and
      Chen, Yun-Nung",
    booktitle = "Findings of the Association for Computational Linguistics: EMNLP 2024",
    month = nov,
    year = "2024",
    address = "Miami, Florida, USA",
    publisher = "Association for Computational Linguistics",
    url = "https://aclanthology.org/2024.findings-emnlp.765/",
    doi = "10.18653/v1/2024.findings-emnlp.765",
    pages = "13078--13096"
}

@article{chao2025natural,
  title={Natural Language Processing and Deep Learning in Cross-Cultural Language Acquisition: From Machine Translation to Cultural Context Understanding},
  author={Chao, Yang},
  journal={Theoretical and Natural Science},
  volume={92},
  pages={13--18},
  year={2025},
  url={https://tns.ewapub.com/article/view/21265}
}
 \clearpage
 \appendix
 \appendixpage
 \addappheadtotoc

\numberwithin{figure}{section}
\numberwithin{table}{section}

 We offer an additional structure as follows:

\begin{itemize}
\item LQM Framework \S\ref{apendx:LQM framework}
\item LQM External Validation \S\ref{appendix:lqm_external_review}
\item Prompts \S\ref{appdx_fig:prompt_generator_no_msa_target}
\item Data Annotation \S\ref{apendx:Data annotation}
\item Linguistic Insights on LLMs Performance \S\ref{apend:Linguistic Insights}
\item Robustness of LQM to Sentence Length \S\ref{sec:lqm_robustness}
\item LQM Fine-Grained Error Distribution  \S\ref{LQM Fine-Grained}
\end{itemize}

\section{\texorpdfstring{LQM Framework}{LQM Framework}}\label{apendx:LQM framework}
Structured along the hierarchy of linguistic analysis, the LQM framework targets six distinct layers: \textbf{sociolinguistic}, \textbf{pragmatic}, \textbf{semantic}, \textbf{morphosyntactic}, \textbf{orthographic/writing conventions}, and \textbf{graphetic}.\\
\noindent\textbf{(i) Sociolinguistic}:
LQM introduces the \textit{code \& register selection} type of error under the sociolinguistic level, explicitly penalizing three main subcategories: \textit{(a)  standardization interference (vertical mismatch)}, where the model reverts to use the standardized, "high," or prestige variety (e.g., Standard German, MSA) when a specific vernacular or "low" variety is requested. For example, the use of MSA appearing in Emirati data as \AR{قوم إبتعد عني} (“Get up, stay away from me”), where \AR{إبتعد عني} is an MSA expression rather than the Emirati one.\\ 
\textit{(b) wrong dialect (horizontal mismatch)}: Where the model uses features specific to a different regional or social variety than the target (e.g., using Mexican Spanish slang in a translation for Spain, or Egyptian idioms in a Levantine text. For example, in the Jordanian sentence \AR{جاسر جا هنا أكتر من مرة، وهيدا ليه مليون سبب.} (“Jasser came here more than once, and this has a million reasons.”), the model mixes Jordanian with other dialects, using the Lebanese form \AR{هيدا} and the Gulf form \AR{جا}.\\
\cite{halliday1978language}'s conceptualization of language as a ``social semiotic,'' wherein register is defined as the specific language variety linked to the field, tenor, and mode of a situation. Accordingly, we include \textit{(c) register mismatch (tone/formality)} as a distinct error type to ensure translations are evaluated not just on propositional content, but on their adherence to the tenor of discourse (e.g., formal vs. informal), a critical dimension often lost in NMT outputs that default to a neutral register (e.g., using "Tu" instead of "Vous" in French, or casual slang in a legal document).\\ 
 
 \textbf{(ii) Pragmatics}:
 While sociolinguistics governs the broad social norms of interaction, the pragmatic level addresses failures in communicative intent and implied meaning, as well as the gap between sentence meaning and speaker meaning. We classify errors at this level under the error types of \textit{use, context, and cultural appropriateness}, capturing instances where the translation is grammatically valid but functionally or culturally anomalous~\cite{farwell1999pragmatics}. In high-context languages like Arabic, literal translation often yields semantically accurate but pragmatically hollow outputs. To capture these nuances, LQM introduces five targeted subcategories:
\textit{(a) speech acts/illocutionary force}: We evaluate whether the translation preserves the \textit{illocutionary force} of the source based on Speech Act Theory \cite{austin1975things, searle1969speech}. NMT models can flatten the pragmatic force of an utterance (e.g., converting a polite request into a direct command). It covers errors in interpreting speech acts (e.g., requests, threats, advice, jokes). For instance, the phrase \AR{ان شاء الله} literally means ``God willing'', but its pragmatic function can vary significantly to mean ``even if'', ``maybe'', or ``hopefully'' depending on use. \\
\textit{(b) code switching}: This category identifies failures where the model either ignores embedded foreign lexical items or forces them into Arabic morphology inappropriately, signaling a failure to recognize the shift in linguistic code.
We categorize CS under pragmatics because language alternation serves specific communicative functions~\cite{gumperz1982discourse}, often includes both situational and metaphorical switching. CS serves as a communicative strategy, linking it to social meaning and interactional goals. for example, the model translated 'I want you to bring me twelve bananas' into Arabic as \AR{ أبيك تجيب لي ١٢ بنانة}. This represents a code-switching error, where the English root is morphologically integrated into an Arabic verb form.\\
\textit{(c) MWEs/proverbs}: Multi-word expressions (MWEs) often function as single non-compositional units. Consistent with theories of semantic non-compositionality~\cite{nunberg1994idioms}, this subcategory targets "literalism errors," where models translate the constituent parts of an idiom rather than its figurative meaning. For instance, the proverb \AR{و مش كل مره تسلم الجره} was translated literally as 'and the jar doesn't stay safe every time.' This fails to convey the underlying warning (i.e., 'you won't get lucky every time'), resulting in a nonsensical output for the target audience.\\
\textit{(d) discourse marker mismatch}:
An inconsistency between the semantic/pragmatic meaning of a discourse marker and the context in which it is used ~\cite{schiffrin1987discourse, fraser2009account}. This can lead to a lack of coherence or an unintended meaning in a conversation or text. Cohesion in Arabic relies on discourse markers (e.g., \textit{wa, fa}) that signal logical relations distinct from English \cite{fraser1999discourse}. We include this subcategory to capture errors where the model misinterprets these procedural cues, disrupting the text's argumentative structure or flow.For example, the model mistranslated the phrase \AR{والله} in the sentence \AR{والله طلبك هو الي هيحدد} as 'God willing, your request is what will determine.' This represents a pragmatic failure; the model interpreted \AR{والله} as a literal religious oath (or confused it with \AR{إن شاء الله}), failing to recognize its function here as a discourse marker used for emphasis/hedging.\\
\textit{(e) vocatives/honorifics/titles}: This category evaluates the translation of forms of address, which serve crucial social functions in communication. We penalize failures to map honorifics or vocative particles (e.g., \textit{Ya}) to their target equivalents, often resulting in a loss of the intended politeness level or social hierarchy \cite{brown1987politeness, mccready2019semantics}. This error frequently occurs when models treat titles as literal nouns rather than pragmatic markers of respect. In this example, the model failed to identify the vocative address in the sentence \AR{أنا أبويه، أحسن عشان النظافة}, mistranslating it as 'I'm his father, it's better for hygiene.' The correct interpretation should be a direct address: 'Dad, I prefer it for the sake of hygiene.\\
\textbf{(iii) Semantics}:
The Semantics level evaluates the fidelity of meaning transfer~\cite{cruse1986lexical}, focusing on the preservation of propositional content. We classify these errors into three primary domains: \textit{(i)} lexical semantics, \textit{(ii)}propositional semantics, and \textit{(iii)} discourse semantics.
\textbf{(a) lexical semantics}: Under this level, we propose subcategories specifically tailored for MT in general, and for Arabic dialectal in particular. These include: \textit{named entity}, failure of the model to translate proper nouns, geographical locations, or organizations correctly. For example, \AR{أم فخري} where (Um Fakhri) is translated into (Um Fakhr).
\textit{wrong term}: This refers to the incorrect use of a term that violates domain-expert usage (e.g., legal or medical, etc.). For instance, the model translates the word comedians'' in the phrase break with comedians is very different'' into \AR{المهرّجين} instead of \AR{ممثلين}; \textit{overtranslation}, where the model uses a specific word (hyponym) when the source used a general word (hypernym) (e.g., “Car” vs. “Ferrari”); \textit{undertranslation}, where the model uses a general word when the source uses a specific word (e.g., “Ferrari” vs. “Vehicle”); \textit{transliteration}: This category applies when the model provides a phonetic rendering of the source text rather than a semantic translation. For example, translating \AR{أيو الحمد لله الحمد لله المهم} as "Ayou al-hamd lillah al-hamd lillah" instead of its English equivalent (e.g., Yes, thank God, anyway''); \textit{unidiomatic/unnatural style}, where the output is grammatically correct but sounds unnatural; \textit{awkward style}, where the style involves excessive wordiness or overly embedded clauses; \textit{unintelligible}, where the output is garbled or incomprehensible; \textit{measurement units}: use of an inappropriate measurement format for its locale; for example, converting units (lbs $\rightarrow$ kg) is a lexical change required to preserve semantic reality; \textit{coverage: unknown term/dialect}: This category includes cases where the model fails to recognize a specific dialectal lemma. For example, in the sentence \AR{سيبوني أكمل يا عواطليه}, the model mistranslated the Egyptian term \AR{عواطليه} ('unemployed') as 'you crazy ones.' This indicates a coverage failure where the model lacked the necessary lexical representation for this dialect-specific term, resulting in incorrect English output; \textit{disambiguation: polysemy failure}, where the model knows the word but selects the wrong meaning (your “Sign” vs “Knock down” example); and \textit{disambiguation: cross-variety interference}, where the model knows the word but assigns it the Standard meaning instead of the Dialect meaning. For example, the model misinterpreted the homograph \AR{شرابين} in the phrase \AR{شرابين حريمي أصفر و أحمر} as 'drinks,' rendering it as 'Two girly yellow and red drinks.' The correct translation is 'two pairs of women's socks.' This represents a failure to distinguish the dialectal term for socks (shurrab) from the standard term for drinks (sharab).\\
\textbf{(b) Propositional Semantics}:
Truth-conditional semantics~\cite{soames1987direct} defines the meaning of a sentence by specifying the conditions under which it is true or false. This category addresses the conservation of information, ensuring that the target text neither adds nor subtracts from the message of the source. We categorize these errors as follows: \textit{addition}, where the translation contains information not found in the source; \textit{omission}, where source content is missing from the target; and \textit{untranslated}, where segments remain in the source language in the final output; and finally \textit{hallucination}: This error occurs when the model generates content that is not supported by the source text, often producing output that is fluent but semantically unrelated. For instance, the model translated \AR{هلأ أختك متأخرها كبير} into 'Now, your sister is very spoiled.' The concept of 'spoiled' is a hallucination; it has no basis in the source text, which actually refers to a 'large delay' or 'arrears' (\AR{متأخرها})\\
\textbf{(c) Discourse Semantics}:
This domain includes four subcategories: \textit{inconsistent use of terminology}, where the model uses multiple terms for the same concept in contexts where consistency is desirable; \textit{inconsistent with terminology resource}, when the model uses a term that differs from term usage required by a specified termbase or other resource; \textit{pronouns}, when the model fails to correctly render pronouns, causing changes in speaker or referent, or flipping the intended gender; and \textit{inconsistent style}, when the style varies inconsistently throughout the text.\\
\textbf{(iv) Morphosyntax}: Evaluates adherence to the structural rules of the target language, focusing on the interface between morphology and syntax \cite{radford2004english}. We classify errors into two types: \textit{grammar}: Addresses violations of morphological agreement \cite{brown2013canonical,corbett2006introduction}, capturing errors in \textit{wrong number, gender, and verb tense}. We include two specific subcategories under this type: \textit{verbal features}, where the translation violates target rules regarding tense, aspect, voice, mood, and person \cite{palmer2001mood}.For example, the Moroccan phrase \AR{نتا دابا عا مخمخ} was mistranslated as 'you’ve really cooked up something.' This indicates a mistranslation of syntactic form, where the model rendered the intended imperative mood as a declarative statement; and \textit{nominal features}, which covers errors in number, gender, case, definiteness, and state (Idafa). For instance, the model failed to translate the subject in this Mauritanian sentence and changed it into something else \AR{يا خياتي راني لكم فاصل} to be "Oh, my brothers, I’m taking a break". \textit{constituent order}: Focuses on the linear arrangement of elements \cite{greenberg1963some}. We explicitly include \textit{address format} and \textit{date format} here, as these require specific reordering to match the target locale conventions rather than direct translation.\\
\clearpage
\begin{table*}[!ht]
\centering
\scriptsize 
\setlength{\tabcolsep}{4pt} 
\renewcommand{\arraystretch}{1.3} 

\begin{tabular}{@{}p{2.0cm}p{3.2cm}p{2.8cm}p{7.5cm}@{}}
\toprule
\rowcolor{gray!20}
\textbf{Category} \newline \tiny(Lightweight LQM) & \textbf{Error Type} \newline \tiny(Lightweight LQM) & \textbf{Subcategory} \newline \textbf{\textit{(Diagnostic LQM)}} & \textbf{Definition} \\
\midrule

\multirowcell{3}[0pt][l]{\cellcolor{catCyan}\textbf{Sociolinguistics}} & \multirowcell{3}[0pt][l]{Language Register} & Standardization Interference  & Use of MSA instead of the target variety. \\
\cmidrule{3-4}
& & Wrong Dialect & Output overlaps with or uses an incorrect dialect. \\
\cmidrule{3-4}
& & Register Mismatch & Formality level higher or lower than required. \\
\midrule

\multirowcell{4}[0pt][l]{\cellcolor{catYellow}\textbf{Pragmatics}} & \multirowcell{4}[0pt][l]{Use, Context, \\ Cultural Appropriateness} & MWEs/ Proverbs & Fails to deliver idiomatic translation; misuse of expression. \\
\cmidrule{3-4}
& & Code Switching & Failure to handle foreign words or recognize code shift. \\
\cmidrule{3-4}
& & Speech Acts/ Illocutionary Force & Intended illocutionary force or speaker intention not conveyed. \\
\cmidrule{3-4}
& & Discourse Marker Mismatch & Misuse of discourse markers affecting cohesion. \\
\midrule

\multirowcell{20}[0pt][l]{\cellcolor{catOrange}\textbf{Semantics}} & \multirowcell{12}[0pt][l]{Lexical Semantics} & Named Entity & Failing to map the proper noun to the correct referent. \\
\cmidrule{3-4}
& & Wrong term & The term is invalid for the domain or creates a conceptual mismatch. \\
\cmidrule{3-4}
& & Overtranslation & Using a specific word ($Hyponym$) for a general source word ($Hypernym$). \\
\cmidrule{3-4}
& & Undertranslation & Using a general word ($Hypernym$) for a specific source word ($Hyponym$). \\
\cmidrule{3-4}
& & Transliteration & Incorrect phonetic rendering into the target language. \\
\cmidrule{3-4}
& & Unidiomatic Style & The style is grammatical but unnatural. \\
\cmidrule{3-4}
& & Awkward Style & Excessive wordiness or overly embedded clauses. \\
\cmidrule{3-4}
& & Unintelligible & The text is garbled or incomprehensible. \\
\cmidrule{3-4}
& & Measurement Units & The measurement format is inappropriate for the locale. \\
\cmidrule{3-4}
& & Coverage & The model fails to recognize a specific dialectal lemma. \\
\cmidrule{3-4}
& & Polysemy Failure & The model picks the wrong meaning for a polysemous word. \\
\cmidrule{3-4}
& & Cross-Variety Interference & Standard meaning assigned instead of the Dialect meaning. \\
\cmidrule{2-4}

& \multirowcell{4}[0pt][l]{Propositional Semantics} & Addition & The target includes information that is not present in the source. \\
\cmidrule{3-4}
& & Omission & Content present in the source is missing from the target. \\
\cmidrule{3-4}
& & Untranslated & The source segment is carried over without translation. \\
\cmidrule{3-4}
& & Hallucination & Adding new information that changes the facts. \\
\cmidrule{2-4}

& \multirowcell{4}[0pt][l]{Discourse Semantics} & Inconsistent use of terminology & Multiple terms used for the same concept. \\
\cmidrule{3-4}
& & Inconsistent with terminology resource & The usage of terms differs from the specified resource. \\
\cmidrule{3-4}
& & Pronouns & Incorrect pronoun causing a change in speaker/gender. \\
\cmidrule{3-4}
& & Inconsistent Style & The style varies inconsistently throughout the text. \\
\midrule

\multirowcell{4}[0pt][l]{\cellcolor{catRed}\textbf{Morphosyntax}} & \multirowcell{2}[0pt][l]{Grammar (wrong number, \\ gender, verb tense)} & Verbal Features & \rule{0pt}{3ex}Violates grammatical rules (Tense, Aspect, Mood, etc). \\
\cmidrule{3-4}
& & Nominal Features & \rule{0pt}{3ex}Violates nominal rules (Number, Gender, Case, etc). \\
\cmidrule{2-4}

& \multirowcell{2}[0pt][l]{constituent order} & Address Format & Inappropriate address format for locale. \\
\cmidrule{3-4}
& & Date format & Inappropriate date format for locale. \\
\midrule

\multirowcell{8}[0pt][l]{\cellcolor{catGreen}\textbf{\shortstack{Orthography/ \\ Writing conventions}}} & \multirowcell{1}[0pt][l]{Spelling} & Typo / Slip & Obvious mechanical error (e.g., typos). \\
\cmidrule{2-4}
& \multirowcell{1}[0pt][l]{Inconsistent Spelling} & - & Same word spelled differently within text. \\
\cmidrule{2-4}
& \multirowcell{1}[0pt][l]{Unconventional Spelling} & - & Spelling hard to read, even if phonetic. \\
\cmidrule{2-4}
& \multirowcell{4}[0pt][l]{Surface Mechanics} & Number Format & Inappropriate number format for locale. \\
\cmidrule{3-4}
& & Currency & Incorrect currency format for locale. \\
\cmidrule{3-4}
& & Time format & Incorrect time format for locale. \\
\cmidrule{3-4}
& & Telephone & Inappropriate telephone number format. \\
\cmidrule{2-4}
& \multirowcell{1}[0pt][l]{Punctuation} & - & Incorrect according to target conventions. \\
\midrule

\multirowcell{1}[0pt][l]{\cellcolor{catMagenta}\textbf{Graphetics}} & - & Character Encoding & Characters garbled due to incorrect encoding. \\

\bottomrule
\end{tabular}
\caption{Hierarchical classification of the LQM. Categories and error types represent the Lightweight LQM; subcategories represent Fine-grained analysis (Diagnostic LQM).}
\label{tab:lqm}
\end{table*}
 \clearpage
\textbf{(v) Orthography/Writing Conventions}: In this level, we evaluate the mechanical correctness of the written output, focusing on the visual representation of language \cite{derwing1992orthographic}. We classify errors into five specific types:
\textit{spelling}: This category addresses deviations from standard orthographic norms. It includes the \textit{typos/slips} subcategory for obvious mechanical errors (e.g., character insertions or deletions) that violate the fixed spelling rules of the target language.
We also distinguish \textit{between inconsistent spelling} and \textit{unconventional spelling}. The category of \textit{surface mechanics} governs non-lexical formatting conventions and comprises four subcategories: \textit{number format}, \textit{currency}, \textit{time format}, and \textit{telephone}.
Finally, \textit{punctuation} addresses cases where punctuation marks are missing, misused, or inconsistent with target language conventions.\\
 \textbf{(vi) Graphetics}: The lowest level of the hierarchy addresses the technical realization of the text code.  We identify one primary failure mode, which is \textit{character encoding}, where the output text is garbled due to incorrect encoding or decoding processes. For example, in this model output \AR{يا أز}$\pm$. Illustrative examples of LQM categories and subcategories are in Table \ref{tab:mqm_full}.
 
\begin{figure}[t]
  \centering
  \includegraphics[width=\linewidth]{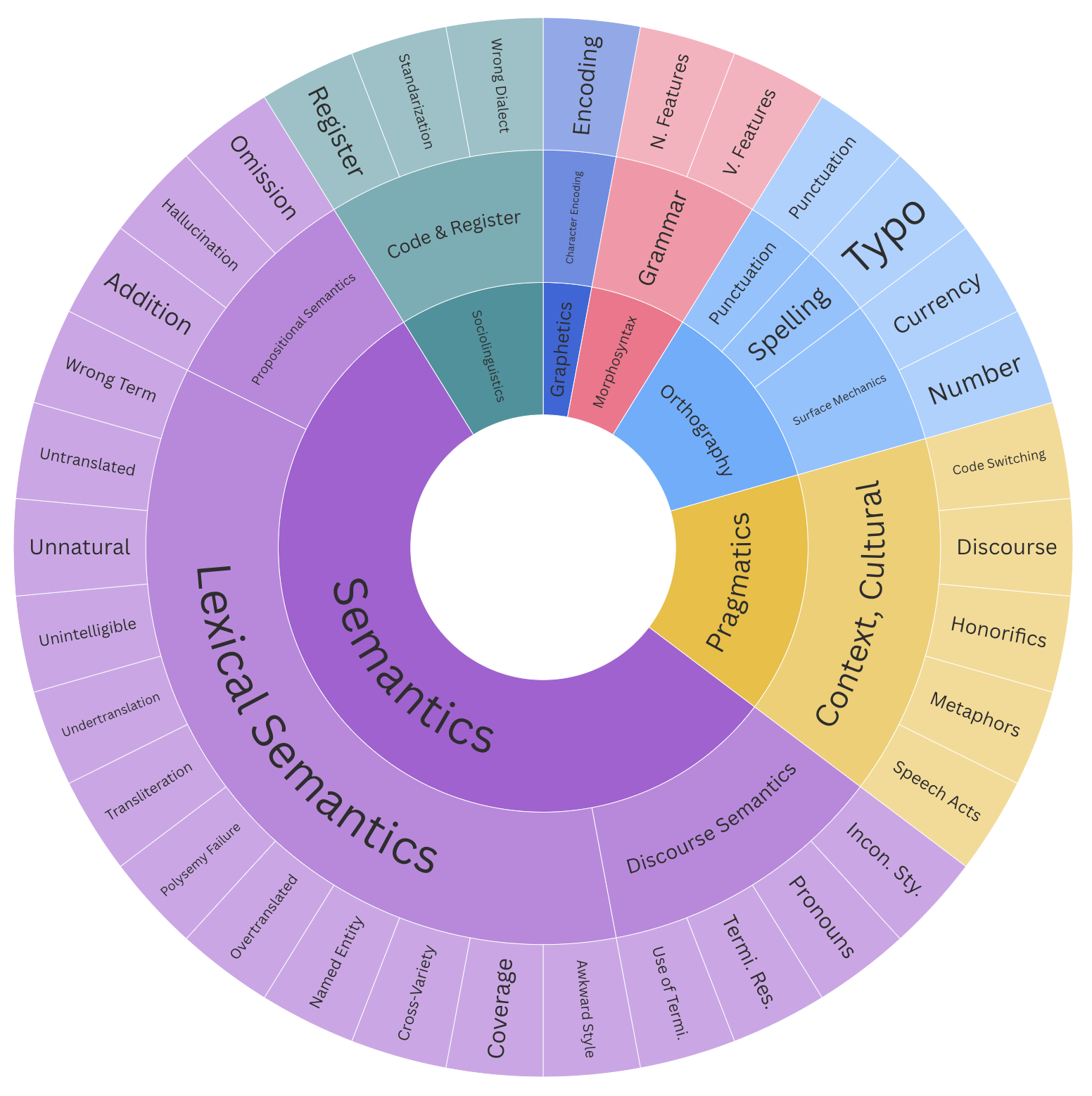}    
 \caption{Distribution of LQM error categories in our dataset, showing that semantic and lexical-semantic errors constitute the majority of labeled errors.}
  \label{Fine-grained-LQM} 
\end{figure}
\section{LQM External Review}
\label{appendix:lqm_external_review}

To obtain external expert feedback on the proposed framework, we invited two linguists specializing in Arabic linguistics and translation to assess its conceptual soundness and practical applicability. Overall, both reviewers viewed the proposed LQM as a valid and promising linguistically motivated framework for machine translation evaluation. In particular, they highlighted its relevance for diglossic language settings, where variation across standard and non-standard varieties, register, and sociolinguistic meaning is central to translation quality. They identified this as a key strength of the framework, especially for Arabic and related contexts.

At the same time, the reviewers noted that some category boundaries, particularly those spanning semantic, pragmatic, and sociolinguistic dimensions, would benefit from further clarification. They also suggested streamlining some fine-grained subcategories to improve annotation consistency and usability. Taken together, their feedback supports the relevance of the framework for MT assessment in diglossic languages while also identifying areas for refinement. In response, we revised the category boundaries to reduce potential overlap and strengthened the definitions and illustrative examples associated with each category.

\section{Prompts}
\label{appdx_fig:prompt_generator_no_msa_target}
\begin{figure*}[ht]
\centering
\begin{tcolorbox}[
    title={\bfseries Prompt},
    width=\textwidth,
    colframe=purple!55!black,
    colback=teal!3,              
    colbacktitle=purple!18,      
    coltitle=black!90,           
    boxsep=3pt,
    arc=1mm,
    fontupper=\small\ttfamily,
    before upper={\setlength{\parskip}{0.5em}\setlength{\baselineskip}{0.9\baselineskip}}
]
{\color{purple!80!black}\bfseries Translation Prompt Generator (Both Directions ENG$\leftrightarrow$Dialect)}

\noindent
def create\_prompt(src, trg):\\
\hspace*{1.5em}if src == 'ENG':\\
\hspace*{3em}prompt = f'Translate the following English phrase into \{langs\_map[trg]\} Arabic dialect written in Arabic script. Your response must only contain the translated text, with no additional explanations or labels.'\\
\hspace*{1.5em}else:\\
\hspace*{3em}prompt = f'Translate the following \{langs\_map[src]\} Arabic dialect phrase into English. Your response must only contain the translated text, with no additional explanations or labels.'\\
\hspace*{1.5em}return prompt

\end{tcolorbox}
\caption{Prompt used to generate all dialectal variants in our dataset (EN→Dialect; Dialect→EN).}
\end{figure*}

\section{\texorpdfstring{Data Annotation}{Data Annotation}}
\label{apendx:Data annotation}
Annotator selection went through multiple stages of quality assurance. First, we developed detailed annotation guidelines that included illustrative examples tailored to the seven Arabic dialects covered in our study. These examples were drawn from a wide range of regional varieties to ensure clarity and consistency for annotators from different Arab countries.
Next, we designed an annotation interface that incorporated all LQM categories along with the subcategories introduced in our framework. We uploaded all model outputs to this interface to facilitate the annotation process. Before launching the full annotation round, we conducted an internal pilot in which we sampled a subset of the data. This allowed us to verify that the tool functioned properly and that all necessary features were available to support accurate and efficient annotation.
During the pilot, we identified several model-generated outputs that were ambiguous or did not fit neatly into the existing error categories. In response, we added a comment box to the interface so that annotators could provide feedback on model behavior or flag unlisted error types, such as pragmatic errors. After refining the tool and guidelines based on these insights, we onboarded all annotators by conducting live demonstrations that walked them through both the annotation procedures and the interface for labeling span-level errors. All annotators were invited to share feedback and comments, which we incorporated as part of our iterative quality assurance process.

\begin{figure*}[h]
\centering
\includegraphics[width=1.0\textwidth]{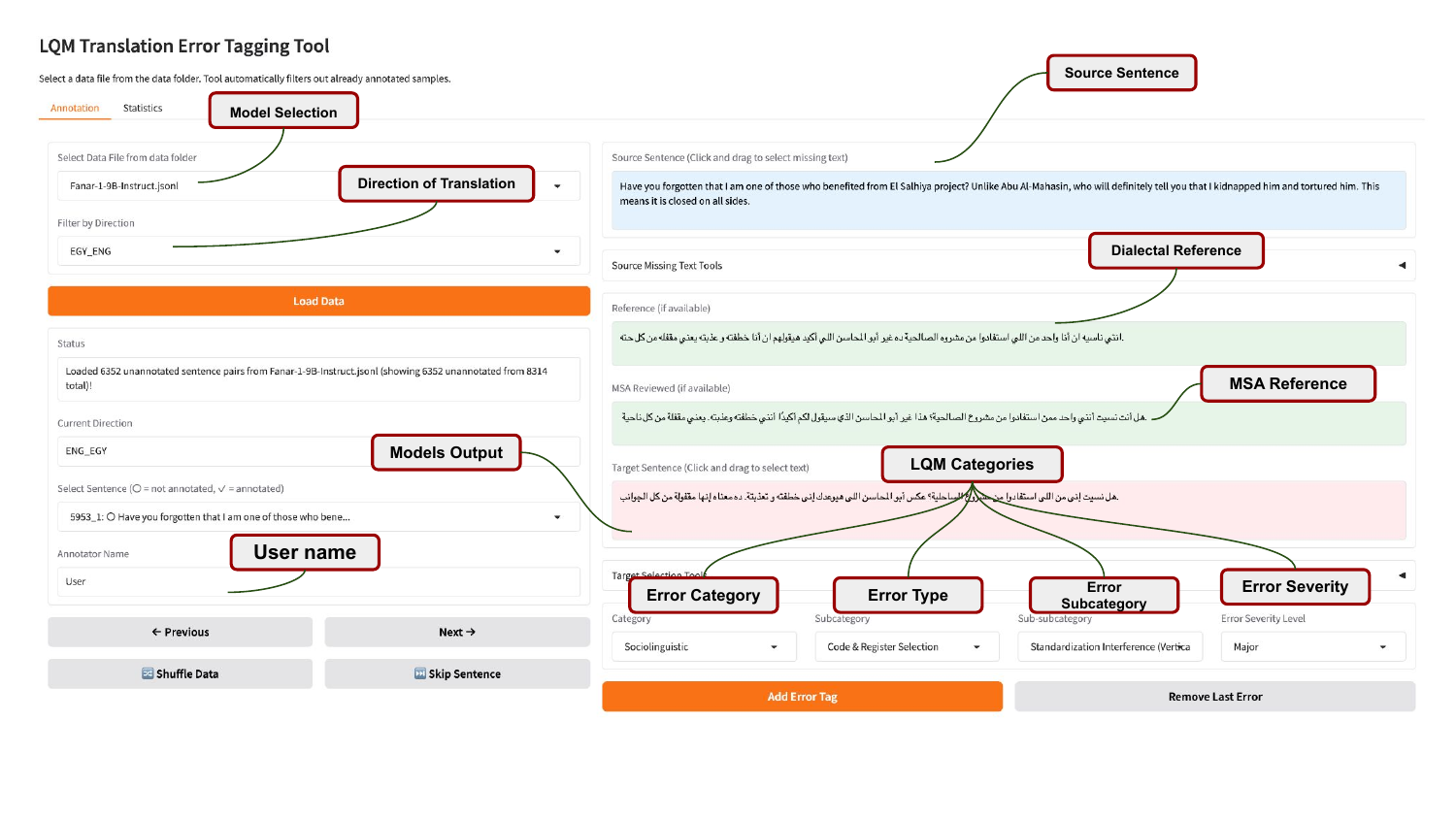}
\caption{Screenshot of our LQM annotation tool for error labeling.}
\label{MQM-platform}
\end{figure*}

\textbf{Annotator Profiles.}
To ensure linguistic expertise and dialectal authenticity, we employed direction-specific annotation teams:
\begin{itemize}[leftmargin=*, itemsep=2pt]
    \item \textbf{Dialect$\rightarrow$English}: This direction was annotated by two senior linguists (Ph.D. holders) specializing in translation studies and linguistics, both of whom are native Egyptian Arabic speakers. To mitigate source-side ambiguity, annotators were provided with both the original dialectal text and its MSA equivalent as a reference. Approximately $40$\% of the dataset was finalized through collaborative live sessions to ensure full agreement, leveraging the MSA context to resolve nuanced dialectal expressions and improve overall comprehension. 
    \item \textbf{English$\rightarrow$Dialect}: This task involved four native speakers (Moroccan, Palestinian, Emirati, and Mauritanian), each holding an MA or Ph.D. in related fields. Each annotator labeled only their native dialect to ensure authentic judgments of naturalness. For the English-to-Egyptian direction, the annotation was performed by the same two linguists mentioned previously.
    For each of the two directions, $40$\% of the data were labeled by two annotators in live sessions with full agreement, once full agreement was reached, the rest of the data was annotated by a single annotator. 
\end{itemize}

\section{Linguistic insights on the LLMs performance}
\label{apend:Linguistic Insights}

The primary limitations of current LLMs in dialectal Arabic--English translation are linguistic rather than merely computational. Our analysis shows that the most persistent errors stem from interpreting dialect-specific semantics, the dominant failure mode in the Dialect $\to$ English direction. Across dialects, Semantic errors account for $67.7$\%--$92.3$\% of total error mass, indicating that the main bottleneck is weak lexical and conceptual mapping for idiomatic and culturally situated expressions. These often encode \textit{illocutionary force} or social alignment that surface-level lexical correspondence cannot recover, producing translations that are formally plausible but pragmatically deficient. This is especially visible in high-resource dialects such as Egyptian, where \textit{Pragmatic} errors reach $24.3$\%. Models also show systematic weakness in dialectal \textit{morphosyntax}; even in comprehension, Morphosyntactic failures remain notable in dialects such as Mauritanian ($3.4$\%), where non-canonical constructions diverge from MSA norms.\\

These challenges intensify and change structurally in the English $\to$ Dialect direction, where failure shifts from semantic decoding to deficient \textit{sociolinguistic} authenticity. Models display a pronounced \textbf{``identity crisis,''} often defaulting to standardization (MSA-vertical mismatch) or producing hybrid forms (wrong dialect-horizontal mismatch). This is reflected in the rise of \textit{Sociolinguistic} errors, peaking at $70.6$\% for UAE and $58.5$\% for Palestinian generation. The problem is especially pronounced in open-source architectures: while \texttt{Pro} contributes as little as $2.1$\% to the error mass in Moroccan generation, open-source models such as \texttt{Fanar} and \texttt{Command-R} contribute up to $33.0$\% and $20.4$\%, respectively. This suggests a data-poverty effect in smaller or open-source models, yielding overgeneralized constructions and fewer culturally grounded pragmatic markers. Overall, these findings show that existing LLMs inadequately model the interaction of morphology, syntax, and sociolinguistic variation, motivating the more granular, linguistically informed evaluation provided by LQM.
\section{Robustness of LQM to Sentence Length}

\label{sec:lqm_robustness}
Per-sentence normalization can inflate or dilute scores for very short or very long segments. To verify that our findings are not artifacts of such effects, we conduct three complementary analyses: \textit{(i)} corpus-level micro-averaged LQM, \textit{(ii)} a robustness check across length buckets, and \textit{(iii)} a rank-stability test using Spearman~$\rho$.


\paragraph{Micro-averaged LQM.}
Instead of averaging per-segment scores, we accumulate all error mass and all token counts across the corpus before dividing once:
\begin{equation}
\mathrm{LQM}_\mu = \max\!\Bigl(0,\; 100 - 100 \cdot \frac{\sum_s E_s}{\sum_s L_s}\Bigr)
\label{eq:micro_lqm}
\end{equation}
This ``sum first, divide once'' principle eliminates the sensitivity to segment length. Table~\ref{tab:lqm_micro_avg} reports micro-averaged scores for selected directions.

\sisetup{
  detect-weight=true,
  detect-family=true,
  round-mode          = places,
  round-precision     = 1,
  table-number-alignment = center
}

\begin{table}[t]
\centering
\small
\setlength{\tabcolsep}{3.5pt}
\resizebox{\columnwidth}{!}{%
\begin{tabular}{
l
S[table-format=2.1]
S[table-format=2.1]
S[table-format=2.1]
S[table-format=2.1]
S[table-format=2.1]
>{\columncolor{blue!15}}S[table-format=2.1]
}
\toprule
{\textbf{Dir.}} & {\textbf{Fanar}} & {\textbf{Cmd-A}} & {\textbf{Cmd-R7B}} & {\textbf{Gemma}} & {\textbf{Flash}} & {\textbf{Pro}} \\
\midrule
Egy$\to$En & 49.4 & 62.5 & 39.7 & 73.0 & 77.4 & \bfseries 77.5 \\
En$\to$Egy & 53.6 & 70.8 & 9.9  & 56.6 & 68.1 & \bfseries 71.9 \\
Pal$\to$En & 67.1 & 78.0 & 64.3 & 74.7 & 76.1 & \bfseries 81.0 \\
Uae$\to$En & 24.0 & 66.2 & 36.8 & 53.2 & 55.3 & \bfseries 66.5 \\
Yem$\to$En & 64.0 & 57.8 & 52.9 & 69.3 & 72.1 & \bfseries 75.0 \\
\midrule
{\textbf{Avg.}} & 51.6 & 67.1 & 40.7 & 65.4 & 69.8 & \bfseries 74.4 \\
\bottomrule
\end{tabular}%
}
\caption{Micro-averaged LQM ($\mathrm{LQM}_{\mu}$) for selected directions. The model ranking is consistent with the per-sentence results in Table~3.}
\label{tab:lqm_micro_avg}
\end{table}

The model ranking under micro-averaged scoring is consistent with the per-sentence results reported in Table~\ref{tab:lqm_scores}. \texttt{Gemini-2.5-Pro} ranks first in the majority of directions under both formulations, and \texttt{Command-R7B} remains the weakest model overall. The direction-level pattern---EN$\to$MOR as the most challenging---is also preserved.


\paragraph{Length-bucket analysis.}
We split all segments into \emph{short}, \emph{medium}, and \emph{long} using the 33rd and 66th percentiles of target-side token count (quantile bucketing ensures balanced counts per bucket). The cut-offs are: \textbf{short}~$\leq 14$ tokens ($n{=}1{,}165$), \textbf{medium}~$15$--$22$ tokens ($n{=}1{,}233$), and \textbf{long}~$> 22$ tokens ($n{=}1{,}080$). We recompute micro-averaged LQM within each bucket for every for each mfor each model--direction combination. Tables~\ref{tab:bucket_egy} and~\ref{tab:bucket_pal} show results for two representative directions. The top-tier models (\texttt{Gemini-2.5-Pro}, \texttt{Gemini-2.5-Flash}) and the weakest model (\texttt{Command-R7B}) maintain their relative positions across all three length buckets in the vast majority of directions.
\begin{table}[t]
\centering
\small

\def\topscore#1{\colorbox{blue!15}{\textbf{#1}}}
\def\secscore#1{\colorbox{blue!5}{#1}}

\setlength{\tabcolsep}{10pt}
\renewcommand{\arraystretch}{1.2} 
\begin{tabular}{lccc} 
\toprule
\rowcolor{blue!8}
\textbf{Model} & \multicolumn{3}{c}{\textbf{Length bucket}} \\
\rowcolor{blue!8}
& \textbf{Short} & \textbf{Medium} & \textbf{Long} \\
\midrule
Gemini-2.5-Pro   & \topscore{66.1} & \topscore{73.5} & \secscore{80.0} \\
Gemini-2.5-Flash & \secscore{66.0} & \secscore{71.1} & \topscore{81.3} \\
\midrule
Gemma-27b        & 58.9            & 65.0            & 78.3 \\
Command-A        & 37.4            & 64.9            & 65.9 \\
Fanar-9B         & 37.8            & 43.2            & 53.9 \\
Command-R7B      & \phantom{0}0.0  & 45.3            & 48.3 \\
\bottomrule
\end{tabular}
\caption{Micro-averaged LQM by length bucket for Egy$\rightarrow$Eng.}
\label{tab:bucket_egy}
\end{table}

\begin{table}[t]
\centering
\small 

\newcommand{\win}[1]{\colorbox{blue!15}{\textbf{#1}}}

\setlength{\tabcolsep}{10pt} 
\renewcommand{\arraystretch}{1.2} 
\begin{tabular}{lccc} 
\toprule
\rowcolor{blue!8}
\textbf{Model} & \textbf{Short} & \textbf{Medium} & \textbf{Long} \\
\midrule
Gemini-2.5-Pro   & \win{76.0} & \win{75.2} & 82.3 \\
Gemma-27b        & 65.6       & 67.2       & 76.8 \\
Gemini-2.5-Flash & 58.7       & 66.8       & 79.5 \\
Command-A        & 50.0       & 60.9       & \win{82.8} \\
Fanar-9B         & 69.8       & 58.3       & 69.7 \\
Command-R7B      & 48.1       & 44.9       & 68.7 \\
\bottomrule
\end{tabular}
\caption{Micro-averaged LQM by length bucket for Pal$\rightarrow$Eng.}
\label{tab:bucket_pal}
\end{table}


\paragraph{Rank stability.}
To quantify stability formally, we compute the Spearman rank correlation of model scores between pairs of length buckets for each direction (Table~\ref{tab:spearman}).
Averaged across all 12 directions, Spearman $\rho = 0.71$ (short vs.\ medium), $0.71$ (medium vs.\ long), and $0.62$ (short vs.\ long), confirming that model rankings are substantially preserved across length strata. The exceptions---Uae$\rightarrow$Eng ($\rho \approx 0.06$--$0.43$) and Eng$\rightarrow$Mau ($\rho \approx -0.15$--$0.70$)---reflect genuine variation in error profiles within those dialect pairs rather than scoring artifacts, consistent with the higher cross-model variability observed for these directions in Table~\ref{tab:lqm_scores}.
\begin{table}[t]
\centering
\footnotesize

\setlength{\tabcolsep}{12pt} 
\renewcommand{\arraystretch}{1.2} 
\begin{tabular}{lccc} 
\toprule
\rowcolor{blue!8}
\textbf{Direction} & \textbf{S\,vs\,M\,$\rho$} & \textbf{M\,vs\,L\,$\rho$} & \textbf{S\,vs\,L\,$\rho$} \\
\midrule
Egy$\rightarrow$Eng & 0.829\rlap{$^{*}$}  & 0.886\rlap{$^{*}$}  & 0.886\rlap{$^{*}$}  \\
Eng$\rightarrow$Egy & 0.943\rlap{$^{**}$} & 0.771               & 0.829\rlap{$^{*}$}  \\
Eng$\rightarrow$Mau & $-$0.696            & $-$0.029            & $-$0.145            \\
Eng$\rightarrow$Mor & 1.000\rlap{$^{**}$} & 0.791               & 0.791               \\
Eng$\rightarrow$Pal & 0.943\rlap{$^{**}$} & 0.886\rlap{$^{*}$}  & 0.771               \\
Eng$\rightarrow$Uae & 0.600               & 0.886\rlap{$^{*}$}  & 0.829\rlap{$^{*}$}  \\
Jor$\rightarrow$Eng & 0.657               & 0.714               & 0.829\rlap{$^{*}$}  \\
Mau$\rightarrow$Eng & 0.829\rlap{$^{*}$}  & 0.886\rlap{$^{*}$}  & 0.886\rlap{$^{*}$}  \\
Mor$\rightarrow$Eng & 0.657               & 0.771               & 0.829\rlap{$^{*}$}  \\
Pal$\rightarrow$Eng & 0.657               & 0.600               & 0.200               \\
Uae$\rightarrow$Eng & 0.145               & 0.429               & 0.058               \\
Yem$\rightarrow$Eng & 0.600               & 0.886\rlap{$^{*}$}  & 0.714               \\
\midrule
\rowcolor{blue!15} 
\textbf{Mean}       & \textbf{0.71}       & \textbf{0.71}       & \textbf{0.62}       \\
\bottomrule
\end{tabular}
\caption{Spearman $\rho$ of model rankings across length buckets. S\,=\,Short, M\,=\,Medium, L\,=\,Long. {$^{*}$\,$p<0.05$}; {$^{**}$\,$p<0.01$}.}
\label{tab:spearman}
\end{table}

\paragraph{Summary.}
Both the micro-averaged formulation and the length-bucket analysis confirm that our main conclusions---\texttt{Gemini-2.5-Pro} leading overall, EN$\to$MOR as the most challenging direction, and direction-dependent shifts in error type---are \textbf{robust to sentence length effects} and are not driven by score distortion in short segments.


\section{LQM Fine-Grained Error Distribution Across Models and Per Direction}
\label{LQM Fine-Grained}
Table~\ref{tab:lqm-fine-grained} shows the error distribution across the $6,113$ samples. It reveals a sharp divide between translation directions, with Dialect-to-English accounting for $58.6$\% of failures compared to $41.4$\% in \textbf{English-to-Dialect}. The most critical generational hurdle in English-to-Dialect is related to \textit{standardization interference}, which represents $35.62$\% of errors in this direction. This indicates a pervasive MSA bias, where models default to formal or prestige varieties rather than maintaining the requested dialectal authenticity. This sociolinguistic failure is further complicated by horizontal dialectal bleed, where features from different regional varieties overlap, leading to a $21.29$\% error rate in dialect target selection (wrong dialect).

In the \textbf{Dialect-to-English direction}, failures are predominantly \textit{semantic} and \textit{pragmatic}. \textit{Named Entity} recognition is the primary bottleneck at $13.49$\%, likely driven by the lack of standardized dialectal orthography, which makes proper noun recovery inconsistent. Furthermore, the model frequently fails to capture the speaker's social intent, with \textit{speech acts} ($2.54$\%) and \textit{vocatives} ($2.49$\%) together accounting for over $5$\% of errors in Dialect-to-English. While these stylistic and lexical issues vary by direction, morphosyntactic struggles—specifically \textit{verbal features}—remain a persistent technical barrier across the board, appearing at significant rates in both decoding ($3.04$\%) and generation ($7.35$\%) tasks.

\begin{table*}[t] 
\centering
\small
\renewcommand{\arraystretch}{1.4} 
\begin{tabular}{clll}
\toprule
\rowcolor{gray!15} & \textbf{Error Sub-category} & \textbf{Dialect $\rightarrow$ Eng (\%)} & \textbf{Eng $\rightarrow$ Dialect (\%)} \\
\midrule
\cellcolor{blue!10} & Standardization Interference & \makebox[24pt][r]{0.00}\,\textcolor{blue!40}{\rule{0.0pt}{6pt}} & \makebox[24pt][r]{35.62}\,\textcolor{teal!40}{\rule{124.7pt}{6pt}} \\
\cellcolor{blue!10} & Wrong Dialect (Horiz. Mismatch) & \makebox[24pt][r]{0.00}\,\textcolor{blue!40}{\rule{0.0pt}{6pt}} & \makebox[24pt][r]{21.29}\,\textcolor{teal!40}{\rule{74.5pt}{6pt}} \\
\multirow{-3.1}{*}{\cellcolor{blue!10}\rotatebox[origin=c]{90}{\textbf{\footnotesize Socioling.}}} & Register Mismatch & \makebox[24pt][r]{0.59}\,\textcolor{blue!40}{\rule{2.1pt}{6pt}} & \makebox[24pt][r]{0.39}\,\textcolor{teal!40}{\rule{1.4pt}{6pt}} \\
\midrule
\cellcolor{green!10} & MWEs/Proverbs & \makebox[24pt][r]{6.28}\,\textcolor{blue!40}{\rule{22.0pt}{6pt}} & \makebox[24pt][r]{2.53}\,\textcolor{teal!40}{\rule{8.9pt}{6pt}} \\
\cellcolor{green!10} & Speech Acts \& Illocutionary Force & \makebox[24pt][r]{2.54}\,\textcolor{blue!40}{\rule{8.9pt}{6pt}} & \makebox[24pt][r]{0.51}\,\textcolor{teal!40}{\rule{1.8pt}{6pt}} \\
\cellcolor{green!10} & Vocatives/Honorifics/Titles & \makebox[24pt][r]{2.49}\,\textcolor{blue!40}{\rule{8.7pt}{6pt}} & \makebox[24pt][r]{0.32}\,\textcolor{teal!40}{\rule{1.1pt}{6pt}} \\
\cellcolor{green!10} & Discourse Marker Mismatch & \makebox[24pt][r]{0.89}\,\textcolor{blue!40}{\rule{3.1pt}{6pt}} & \makebox[24pt][r]{0.12}\,\textcolor{teal!40}{\rule{0.4pt}{6pt}} \\
\multirow{-5.1}{*}{\cellcolor{green!10}\rotatebox[origin=c]{90}{\textbf{\footnotesize Pragmatic}}} & Code Switching & \makebox[24pt][r]{0.11}\,\textcolor{blue!40}{\rule{0.4pt}{6pt}} & \makebox[24pt][r]{0.59}\,\textcolor{teal!40}{\rule{2.1pt}{6pt}} \\
\midrule
\cellcolor{yellow!15} & NE (Named Entities) & \makebox[24pt][r]{13.49}\,\textcolor{blue!40}{\rule{47.2pt}{6pt}} & \makebox[24pt][r]{3.52}\,\textcolor{teal!40}{\rule{12.3pt}{6pt}} \\
\cellcolor{yellow!15} & Coverage: Unknown Term/Dialect & \makebox[24pt][r]{9.33}\,\textcolor{blue!40}{\rule{32.7pt}{6pt}} & \makebox[24pt][r]{1.22}\,\textcolor{teal!40}{\rule{4.3pt}{6pt}} \\
\cellcolor{yellow!15} & Omission & \makebox[24pt][r]{8.94}\,\textcolor{blue!40}{\rule{31.3pt}{6pt}} & \makebox[24pt][r]{1.46}\,\textcolor{teal!40}{\rule{5.1pt}{6pt}} \\
\cellcolor{yellow!15} & Unnatural/Unidiomatic Style & \makebox[24pt][r]{6.65}\,\textcolor{blue!40}{\rule{23.3pt}{6pt}} & \makebox[24pt][r]{3.32}\,\textcolor{teal!40}{\rule{11.6pt}{6pt}} \\
\cellcolor{yellow!15} & Awkward Style & \makebox[24pt][r]{4.61}\,\textcolor{blue!40}{\rule{16.1pt}{6pt}} & \makebox[24pt][r]{5.02}\,\textcolor{teal!40}{\rule{17.6pt}{6pt}} \\
\cellcolor{yellow!15} & Addition & \makebox[24pt][r]{5.86}\,\textcolor{blue!40}{\rule{20.5pt}{6pt}} & \makebox[24pt][r]{2.09}\,\textcolor{teal!40}{\rule{7.3pt}{6pt}} \\
\cellcolor{yellow!15} & Pronouns & \makebox[24pt][r]{6.84}\,\textcolor{blue!40}{\rule{23.9pt}{6pt}} & \makebox[24pt][r]{0.91}\,\textcolor{teal!40}{\rule{3.2pt}{6pt}} \\
\cellcolor{yellow!15} & Disambiguation: Cross-Variety & \makebox[24pt][r]{6.67}\,\textcolor{blue!40}{\rule{23.3pt}{6pt}} & \makebox[24pt][r]{0.28}\,\textcolor{teal!40}{\rule{1.0pt}{6pt}} \\
\cellcolor{yellow!15} & Wrong Term & \makebox[24pt][r]{3.18}\,\textcolor{blue!40}{\rule{11.1pt}{6pt}} & \makebox[24pt][r]{3.08}\,\textcolor{teal!40}{\rule{10.8pt}{6pt}} \\
\cellcolor{yellow!15} & Hallucination & \makebox[24pt][r]{4.47}\,\textcolor{blue!40}{\rule{15.6pt}{6pt}} & \makebox[24pt][r]{1.18}\,\textcolor{teal!40}{\rule{4.1pt}{6pt}} \\
\cellcolor{yellow!15} & Undertranslation & \makebox[24pt][r]{2.90}\,\textcolor{blue!40}{\rule{10.2pt}{6pt}} & \makebox[24pt][r]{1.03}\,\textcolor{teal!40}{\rule{3.6pt}{6pt}} \\
\cellcolor{yellow!15} & Disambiguation: Polysemy Failure & \makebox[24pt][r]{3.02}\,\textcolor{blue!40}{\rule{10.6pt}{6pt}} & \makebox[24pt][r]{0.47}\,\textcolor{teal!40}{\rule{1.6pt}{6pt}} \\
\cellcolor{yellow!15} & Transliteration & \makebox[24pt][r]{2.96}\,\textcolor{blue!40}{\rule{10.4pt}{6pt}} & \makebox[24pt][r]{0.36}\,\textcolor{teal!40}{\rule{1.3pt}{6pt}} \\
\cellcolor{yellow!15} & Overtranslation & \makebox[24pt][r]{1.65}\,\textcolor{blue!40}{\rule{5.8pt}{6pt}} & \makebox[24pt][r]{0.28}\,\textcolor{teal!40}{\rule{1.0pt}{6pt}} \\
\cellcolor{yellow!15} & Untranslated & \makebox[24pt][r]{1.42}\,\textcolor{blue!40}{\rule{5.0pt}{6pt}} & \makebox[24pt][r]{0.24}\,\textcolor{teal!40}{\rule{0.8pt}{6pt}} \\
\cellcolor{yellow!15} & Inconsistent Term. Resource & \makebox[24pt][r]{1.23}\,\textcolor{blue!40}{\rule{4.3pt}{6pt}} & \makebox[24pt][r]{0.12}\,\textcolor{teal!40}{\rule{0.4pt}{6pt}} \\
\cellcolor{yellow!15} & Inconsistent Style & \makebox[24pt][r]{0.03}\,\textcolor{blue!40}{\rule{0.1pt}{6pt}} & \makebox[24pt][r]{0.55}\,\textcolor{teal!40}{\rule{1.9pt}{6pt}} \\
\cellcolor{yellow!15} & Unintelligible & \makebox[24pt][r]{0.03}\,\textcolor{blue!40}{\rule{0.1pt}{6pt}} & \makebox[24pt][r]{0.16}\,\textcolor{teal!40}{\rule{0.6pt}{6pt}} \\
\multirow{-19.1}{*}{\cellcolor{yellow!15}\rotatebox[origin=c]{90}{\textbf{\footnotesize Semantic}}} & Inconsistent Use of Terminology & \makebox[24pt][r]{0.03}\,\textcolor{blue!40}{\rule{0.1pt}{6pt}} & \makebox[24pt][r]{0.00}\,\textcolor{teal!40}{\rule{0.0pt}{6pt}} \\
\midrule
\cellcolor{purple!10} & Verbal Features & \makebox[24pt][r]{3.04}\,\textcolor{blue!40}{\rule{10.6pt}{6pt}} & \makebox[24pt][r]{7.35}\,\textcolor{teal!40}{\rule{25.7pt}{6pt}} \\
\multirow{-2.1}{*}{\cellcolor{purple!10}\rotatebox[origin=c]{90}{\textbf{\footnotesize Morph.}}} & Nominal Features & \makebox[24pt][r]{0.28}\,\textcolor{blue!40}{\rule{1.0pt}{6pt}} & \makebox[24pt][r]{0.79}\,\textcolor{teal!40}{\rule{2.8pt}{6pt}} \\
\midrule
\cellcolor{orange!15} & Typo/Slip & \makebox[24pt][r]{0.14}\,\textcolor{blue!40}{\rule{0.5pt}{6pt}} & \makebox[24pt][r]{4.74}\,\textcolor{teal!40}{\rule{16.6pt}{6pt}} \\
\cellcolor{orange!15} & Currency & \makebox[24pt][r]{0.14}\,\textcolor{blue!40}{\rule{0.5pt}{6pt}} & \makebox[24pt][r]{0.16}\,\textcolor{teal!40}{\rule{0.6pt}{6pt}} \\
\cellcolor{orange!15} & Punctuation & \makebox[24pt][r]{0.08}\,\textcolor{blue!40}{\rule{0.3pt}{6pt}} & \makebox[24pt][r]{0.20}\,\textcolor{teal!40}{\rule{0.7pt}{6pt}} \\
\multirow{-4.1}{*}{\cellcolor{orange!15}\rotatebox[origin=c]{90}{\textbf{\footnotesize Ortho.}}} & Number Format & \makebox[24pt][r]{0.00}\,\textcolor{blue!40}{\rule{0.0pt}{6pt}} & \makebox[24pt][r]{0.04}\,\textcolor{teal!40}{\rule{0.1pt}{6pt}} \\
\midrule
\cellcolor{red!10}\rotatebox[origin=c]{90}{\textbf{\footnotesize Graph.}} & Character Encoding & \makebox[24pt][r]{0.06}\,\textcolor{blue!40}{\rule{0.2pt}{6pt}} & \makebox[24pt][r]{0.08}\,\textcolor{teal!40}{\rule{0.3pt}{6pt}} \\
\bottomrule
\end{tabular}
\caption{Fine-grained LQM error distribution. Bar lengths are proportional to the percentage of total errors per translation direction.}
\label{tab:lqm-fine-grained}
\end{table*}

\clearpage
{
\scriptsize
\setlength{\tabcolsep}{1.5pt}
\setlength{\fboxsep}{1pt}
\renewcommand{\arraystretch}{1.35}
\onecolumn

\setlength{\LTleft}{0pt}
\setlength{\LTright}{0pt}

\begin{longtable}{@{}
    >{\RaggedRight\hspace{0pt}}m{1.55cm}
    >{\RaggedRight\hspace{0pt}}m{1.95cm}
    >{\RaggedRight\hspace{0pt}}m{1.95cm}
    >{\centering\arraybackslash}m{1.15cm} 
    >{\RaggedRight\hspace{0pt}}m{3.80cm}  
    >{\RaggedRight\hspace{0pt}}m{3.80cm}  
    >{\RaggedRight\hspace{0pt}}m{1.55cm}
@{}}

 \\
\toprule
\rowcolor{TblHeaderBG}
\textcolor{TblHeaderFG}{\textbf{Category}} &
\textcolor{TblHeaderFG}{\textbf{Error Type}} &
\textcolor{TblHeaderFG}{\textbf{Subcategory}} &
\textcolor{TblHeaderFG}{\textbf{Dir}} &
\textcolor{TblHeaderFG}{\textbf{Source}} &
\textcolor{TblHeaderFG}{\textbf{Target}} &
\textcolor{TblHeaderFG}{\textbf{Model}} \\
\midrule
\endfirsthead

\multicolumn{7}{c}{{\bfseries \tablename\ \thetable{} -- continued from previous page}} \\
\toprule
\rowcolor{TblHeaderBG}
\textcolor{TblHeaderFG}{\textbf{Category}} &
\textcolor{TblHeaderFG}{\textbf{Error Type}} &
\textcolor{TblHeaderFG}{\textbf{Subcategory}} &
\textcolor{TblHeaderFG}{\textbf{Dir}} &
\textcolor{TblHeaderFG}{\textbf{Source}} &
\textcolor{TblHeaderFG}{\textbf{Target}} &
\textcolor{TblHeaderFG}{\textbf{Model}} \\
\midrule
\endhead

\bottomrule
\multicolumn{7}{r}{{Continued on next page...}} \\
\endfoot

\bottomrule
\endlastfoot

\multirow{3}{*}{\mcell{1.55cm}{\textbf{Sociolinguistics}}} &
\multirow{3}{*}{\mcell{1.95cm}{Code \& Register Selection}} &
Standardization Interference &
\DirPill{ENG}{MOR} &
My phone was silent so I did not see your calls. &
\AR{تلفوني كان }\hlerror{\AR{صامت}}\AR{ باش ما شفتش ديالك ديال المكالمات.} &
\ModelBadge{GemmaC}{$\blacklozenge$}{gemma-3} \\
 &  &
Wrong Dialect &
\DirPill{UAE}{ENG} &
May Allah protect you. You are good and blessed. &
\AR{الله يحفظك. إنت }\hlerror{\AR{زين}}\AR{ ومبارك. } &
\ModelBadge{GeminiC}{$\bigstar$}{Gemini-F} \\
 &  &
Register Mismatch &
\DirPill{EN}{MOR} &
What are you doing, man?  &
\AR{آش كادير }\hlerror{\AR{أ بنادم}}\AR{ } &
\ModelBadge{GeminiC}{$\bigstar$}{gemini-2.5-flash} \\
\midrule

\multirow{5}{*}{\mcell{1.55cm}{\textbf{Pragmatics}}} &
\multirow{5}{*}{\mcell{1.95cm}{Use, Context, Cultural Approp}} &
MWEs/ Proverbs &
\DirPill{EGY}{ENG} &
\AR{ بقولك ايه ما تيجي ناكل لقمه سوا عشان يبقى عيش و ملح؟} &
Tell you what, why don't we grab a bite together so there's \hlerror{bread and salt between us}? &
\ModelBadge{GeminiC}{$\bigstar$}{gemini-2.5-pro} \\
 &  &
Code Switching &
\DirPill{ENG}{EGY} &
It's not called a casino, it's called a nightclub, ma'am. &
\AR{مش اسمه كازينو، اسمه نايت كلوب يا مدام.} &
\ModelBadge{GeminiC}{$\bigstar$}{gemini-2.5-flash} \\
 &  &
Speech Acts/ Illoc Force &
\DirPill{MOR}{ENG} &
\AR{ياك؟ دابا نتا مشيتي و جمعتي و سمعتي هاد التخربيق.} &
Right? Now you went, gathered, and listened to this nonsense.  &
\ModelBadge{GemmaC}{$\blacklozenge$}{gemma-3} \\
 &  &
Discourse Marker Mismatch &
\DirPill{MOR}{ENG} &
\AR{الصراحة بابا انا شدني لو تخاك و تبلوكيت اوغوزمون مهدي عتق الموقف} &
To be honest, dad, I’m really embarrassed about what happened with that girl and the photos, and \hlerror{honestly}, Mehdi saved the situation. &
\ModelBadge{GemmaC}{$\blacklozenge$}{gemma-3} \\
 &  &
Vocatives /Honorifics/Titles &
\DirPill{YEM}{ENG} &
\AR{يعني ايش يا اباه ، خلاص عاتجسمو ، ونورث ، وتوجع امورنا سابره.} &
So what, \hlerror{man}, just toughen up, inherit [this], and make our affairs difficult. &
\ModelBadge{GemmaC}{$\blacklozenge$}{gemma-3} \\
\midrule

\multirow{15}{*}{\mcell{1.55cm}{\textbf{Semantics}}} &
\multirow{10}{*}{\mcell{1.95cm}{Lexical Semantics}} &
Named Entity &
\DirPill{ENG}{PAL} &
\AR{والله ذمتك وسيعة يا ابو عطا يا دكر! } &
By God, your heart is wide, Abu Atta, you \hlerror{rooster}! &
\ModelBadge{FanarC}{$\blacktriangle$}{Fanar-1-9B} \\
 &  &
Wrong term  &
\DirPill{ENG}{PAL} &
\AR{والله لو بحجوا مرة و مرتين و ثلاث و بضلوا يسعوا و يطوفوا و يسعوا و يطوفوا ما بنغفرلهم ذنب بعد اللي عملوه في ولاد البلد} &
by God, even if they go for Hajj once, twice, three times, and keep \hlerror{circumambulating and walking}, we won't forgive them for their sin after what they did to the children of the town. &
\ModelBadge{FanarC}{$\blacktriangle$}{Fanar-1-9B} \\
 &  &
Overtranslation &
\DirPill{MAU}{ENG} &
\AR{ذا اشوي عايشين منو وكارين منو دار وذا كامل طامع انت فينا ظرك نعطوك عشرين ألف} &
You see how we live, we are barely surviving, and you want everything, you \hlerror{think} we will give you twenty thousand? &
\ModelBadge{FanarC}{$\blacktriangle$}{Fanar-1-9B} \\
 &  &
Undertranslation &
\DirPill{YEM}{ENG} &
\AR{خليها على الله بس, امانه ان قلبي مغدغد غدغده عليه} &
Leave it to God, but honestly, my heart is \hlerror{restless with worry} about it. &
\ModelBadge{FanarC}{$\blacktriangle$}{Fanar-1-9B} \\
 &  &
Transliteration &
\DirPill{EGY}{ENG} &
\AR{سعد عارف ناصف مدكور زي ما حنا عارفينه بالزبط} &
Saad \hlerror{Arafa} Nasif Madkor, as we all know. &
\ModelBadge{CommandC}{$\blacktriangleright$}{c4ai-command-r7b} \\
 &  &
Unnatural Style  &
\DirPill{MOR}{ENG} &
\AR{المرة الجاية غا ندير فحالك غاندير راحة بحالا سخفت.} &
Next time I'll do like you, I'll \hlerror{take it easy} like I'm weak or tired. &
\ModelBadge{CommandC}{$\blacktriangleright$}{c4ai-command-r7b} \\
 &  &
Awkward Style &
\DirPill{PAL}{ENG} &
\AR{يا سلام يا سلام كل البلد ورا و سلامة قدام! بس انا بقيت مشغول منت عارف} &
\hlerror{Peace, peace}, the whole country is behind and safety is in front! But I've become busy, I don't know. &
\ModelBadge{CommandC}{$\blacktriangleright$}{c4ai-command-r7b} \\
 &  &
Unknown Term/ Dialect &
\DirPill{UAE}{ENG} &
\AR{شو كان ييب حق المتوه ؟، و عيد شو كان ييب حق المتوه ؟} &
What did he bring for \hlerror{the guest}? What did he bring for the guest? &
\ModelBadge{CommandC}{$\blacktriangleright$}{c4ai-command-r7b} \\
 &  &
Polysemy Failure &
\DirPill{JOR}{ENG} &
\AR{هات جاي بوقع لك و بوقع لعشرة مثلك كمان شو يعني فكرك أنا خويف يعني؟} &
Give it to me, I will sign it for you and ten people like you. Do you think I am afraid? &
\ModelBadge{GemmaC}{$\blacklozenge$}{gemma-3} \\
 &  &
Cross-Variety Interference &
\DirPill{JOR}{ENG} &
\AR{ديري بالك على الصندوق يا آنسه ليلى بدي كل شي مظبوط و دقيق} &
Take care of \hlerror{the box}, Miss Leila. I want everything to be accurate and precise. &
\ModelBadge{CommandC}{$\blacktriangleright$}{c4ai-command-a-03} \\

\cmidrule(lr){2-7}
 & \multirow{4}{*}{\mcell{1.95cm}{Propositional Semantics}} &
Addition  &
\DirPill{MAU}{ENG} &
\AR{! آن سياني عندي جيتو من امو جيتو من خوه جيتو من ظهرو } &
I have a problem with my mother, my brother, and my \hlerror{father}. &
\ModelBadge{FanarC}{$\blacktriangle$}{Fanar-1-9B} \\
 &  &
Omission &
\DirPill{UAE}{ENG} &
\AR{زين }\hlerror{\AR{بويه}}\AR{ وين حرمنا المصون زقروها ليي مشتاق ليها، سيري زقريها عمو } &
but where is our private space? I miss it, go, show it to me, uncle. &
\ModelBadge{FanarC}{$\blacktriangle$}{Fanar-1-9B} \\
 &  &
Untranslated  &
\DirPill{JOR}{ENG} &
\AR{طيب يا يوسف يعني خليها لبكرة الصبح.} &
\AR{Okay, }\hlerror{\AR{يوسف}}\AR{, let's do it tomorrow morning.} &
\ModelBadge{FanarC}{$\blacktriangle$}{Fanar-1-9B} \\
 &  &
Hallucination &
\DirPill{EGY}{ENG} &
\AR{أنا برضو انتي مش كنتي انتي اللي قاعدة .} &
\hlerror{I also thought you weren’t the one sitting there}, you stupid woman.  &
\ModelBadge{CommandC}{$\blacktriangleright$}{c4ai-command-a} \\

\cmidrule(lr){2-7}
 & Discourse Semantics &
Pronouns  &
\DirPill{MAU}{ENG} &
\AR{شوف خليك لي ذلي قلت لك ول ذلي ليهي تقول لي جوابي فيه.} &
Look, stay with me, I told you, and stay with her &
\ModelBadge{CommandC}{$\blacktriangleright$}{c4ai-command-a} \\
\midrule

\multirow{2}{*}{\mcell{1.55cm}{\textbf{Morphosyntax}}} &
\multirow{2}{*}{\mcell{1.95cm}{Grammar (number, gender, tense)}} &
Verbal Features &
\DirPill{MOR}{ENG} &
\AR{شوف ا موحا سير الله يرضي عليك عند دوك الناس لي كا يصاوبولنا هاد الحويجات } &
may God be pleased with you. \hlerror{Look, there are people who} are helping us with these needs &
\ModelBadge{FanarC}{$\blacktriangle$}{Fanar-1-9B} \\
 &  &
Nominal Features &
\DirPill{ENG}{EGY} &
I came to check on you, my love, because we heard women’s feet going up the stairs. I was afraid about you, Kuka. &
\AR{جيت أطمن عليكي يا حبيبتي، عشان سمعنا }\hlerror{\AR{رجلي}}\AR{ حريم طالعة على السلم. كنت خايفة عليكي يا كوكا.} &
\ModelBadge{CommandC}{$\blacktriangleright$}{c4ai-command-a} \\
\midrule

\multirow{2}{*}{\mcell{1.55cm}{\textbf{Orthography/ Writing conventions}}} &
Unconventional Spelling &
 &
\DirPill{ENG}{PAL} &
What do I know, uncle? Is Ma’rouf the imposter better than us!? &
\AR{شو بتعرف أنا، يا عمّي؟ معقول }\hlerror{\AR{الدّعيّ}}\AR{ معرووف أحسن منّا؟!} &
\ModelBadge{GemmaC}{$\blacklozenge$}{gemma-3} \\
\cmidrule(lr){2-7}
 & Surface Mechanics &
Currency &
\DirPill{ENG}{UAE} &
The fellow will give me ten rubies, father. I told you I went and talked to him, and now she has agreed. &
\AR{الرجال بيعطيني عشر }\hlerror{\AR{ياقوت}}\AR{، يبه. قلت لك رحت وكلمته، و الحين وافقت.} &
\ModelBadge{GemmaC}{$\blacklozenge$}{gemma-3-27b} \\ 
\bottomrule

\noalign{\smallskip}
\caption{LQM error types across all linguistic levels examples from covered dialects.} \label{tab:mqm_full} 
\end{longtable}
}





\appendix

\end{document}